\documentclass[robotics,article,submit,pdftex,moreauthors]{Definitions/mdpi} 
\usepackage{graphics} %
\usepackage{epsfig} %
\usepackage{amsmath} %
\usepackage{amssymb}  %
\usepackage{amsthm}
\usepackage{upgreek}
\usepackage[utf8]{inputenc}
\usepackage{xcolor}

\usepackage{graphicx}

\usepackage{amsmath}
\usepackage{stfloats}

\usepackage[official]{eurosym}

\usepackage{algorithm}
\usepackage{algpseudocode}
\usepackage{epstopdf}
\usepackage[leftcaption]{sidecap}
\usepackage{booktabs}
\usepackage{siunitx}
\usepackage{multirow}

\usepackage{glossaries}

\newacronym{hyq}{HyQ}{Hydraulically actuated Quadruped}

\newacronym{lf}{LF}{Left-Front}
\newacronym{rf}{RF}{Right-Front}
\newacronym{lh}{LH}{Left-Hind}
\newacronym{rh}{RH}{Right-Hind}

\newacronym{haa}{HAA}{Hip Adduction-Abduction}
\newacronym{hfe}{HFE}{Hip Flexion-Extension}
\newacronym{kfe}{KFE}{Knee Flexion-Extension}

\newacronym{imu}{IMU}{Inertial Measurement Unit}
\newacronym{dofs}{DoFs}{Degrees of Freedom}
\newacronym{rt}{RT}{Real Time}

\newacronym{com}{CoM}{Center of Mass}
\newacronym{cop}{CoP}{Center of Pressure}
\newacronym{zmp}{ZMP}{Zero Moment Point}
\newacronym{icp}{ICP}{Instantaneous Capture Point}
\newacronym{cmp}{CMP}{Centroidal Moment Pivot}
\newacronym{grfs}{GRFs}{Ground Reaction Forces}

\newacronym{ls}{LS}{Least Square}

\newacronym{slip}{SLIP}{Spring Loaded Inverted Pendulum}
\newacronym{eom}{EoM}{Equation of Motions}
\newacronym{qp}{QP}{Quadratic Program}
\newacronym{sqp}{SQP}{Sequential Quadratic Programming}
\newacronym{mic}{MIC}{Mixed-Integer Convex}
\newacronym{cmaes}{CMA-ES}{Covariance Matrix Adaptation Evolution Strategy}
\newacronym{ara}{ARA*}{Anytime Repairing A*}
\newacronym{pca}{PCA}{Principal Component Analysis}
\newacronym{cpg}{CPG}{Central Pattern Generator}
\newacronym{wbc}{WBC}{Whole-Body Control}

\newacronym{pd}{PD}{Proportional-Derivative}

\newacronym{mpc}{MPC}{Model Predictive Control}
\newacronym{nmpc}{NMPC}{Nonlinear Model Predictive Control}
\newacronym{awbc}{c$^3$WBC}{Compliant Contact Consistent Whole-Body Control}
\newacronym{swbc}{sWBC}{Standard Whole-Body Control}
\newacronym{c3wbc}{c$^3$WBC}{Compliant Contact Consistent Whole-Body Control}
\newacronym{ste}{TCE}{Terrain Compliance Estimator}
\newacronym{c3}{\texttt{c}$^3$}{compliant contact consistent}

\newacronym{stance}{STANCE}{\textbf{S}oft \textbf{T}errain \textbf{A}daptation a\textbf{n}d \textbf{C}ompliance \textbf{E}stimation}

\newacronym{wbopt}{WBOpt}{Whole Body Optimization}

\newacronym{hc}{HC}{Hunt and Crossley's}
\newacronym{kv}{KV}{Kelvin-Voigt's}

\newacronym{wllsr}{WLLSR}{Weighted Linear Least Squared Regression}

\newacronym{mae}{MAE}{Mean Absolute Tracking Error}

\newacronym{ode}{ODE}{Open Dynamics Engine}
\newacronym{lip}{LIP}{Linear Inverted Pendulum}
\newacronym{srbd}{SRBD}{Single Rigid Body Dynamics}

\newcommand{\Rnum}{\mathbb{R}} %

\newcommand{\vect}[1]{\mathbf{#1}} %
\newcommand{\bm}[1]{\boldsymbol{#1}}

\newcommand{\mx}[1]{\mathbf{\bm{#1}}} 				%

\newcommand{\mat}[1]{\ensuremath{\begin{bmatrix}#1\end{bmatrix}}}	%
\newcommand{\x}{\ensuremath{\times}}

\firstpage{1} 
\allowdisplaybreaks
\pubvolume{1}
\issuenum{1}
\articlenumber{0}
\pubyear{2023}
\copyrightyear{2023}
\datereceived{} 
\dateaccepted{} 
\datepublished{} 
\hreflink{https://doi.org/} %
\pdfoutput=1

\Title{Optimization-Based Reference Generator for Nonlinear Model Predictive Control of Legged Robots}

\TitleCitation{Optimization-Based Reference Generator for Nonlinear Model Predictive Control of Legged Robots}

\Author{Angelo Bratta$^{1,2,*}$, Michele Focchi$^{1,3,*}$, Niraj Rathod$^{1,4}$, and Claudio Semini$^{1}$}

\AuthorNames{Angelo Bratta, Michele Focchi, Niraj Rathod, Mario Zanon, Alberto Bemporad and Claudio Semini}

\AuthorCitation{A. Bratta, M. Focchi, N. Rathod, C. Semini}

\address{%
$^{1}$ \quad Dynamic Legged Systems Lab, Istituto Italiano di Tecnologia (IIT), Genova, Italy\\
$^{2}$ \quad Dipartimento di Informatica, Bioingegneria, Robotica e Ingegneria dei Sistemi (DIBRIS), Università di Genova, Genova, Italy\\
$^{3}$ \quad Dipartimento di Ingegneria e Scienza dell'Informazione (DISI), Università di Trento, Trento, Italy\\
$^{4}$ \quad IMT School for Advanced Studies Lucca, Lucca, Italy}

\corres{Corresponding authors: angelo.bratta@iit.it, michele.focchi@iit.it}

\abstract{Model Predictive Control (MPC) approaches are widely used in robotics, since they guarantee feasibility and allow the computation of updated trajectories while the robot is moving.
	They generally require heuristic references for the tracking terms and proper tuning of the parameters of the cost function in order to obtain good performance. 
	For instance, when a legged robot has to react to disturbances from the environment (e.g., recover after a push) or track a specific goal with statically unstable gaits, the effectiveness of the algorithm can degrade.
	In this work, we propose a novel optimization-based Reference Generator which exploits a Linear Inverted Pendulum (LIP) model to 
	compute reference trajectories for the Center of Mass while taking into account the possible under-actuation of a gait (e.g., in a trot).
	The obtained trajectories are used as references for the cost function of the Nonlinear MPC presented in our previous work \cite{openaccess}. 
	We also present a formulation that ensures guarantees on the response time to reach a goal without the need to tune the weights of the cost terms. 
	In addition, footholds are corrected using the optimized reference
	to drive the robot towards the goal.
	We demonstrate the effectiveness of our approach both in simulations and experiments in different scenarios with the Aliengo robot.}

\begin{document}
\section*{Notation}
Most commonly used symbols in this article.\\

\begin{tabular}{@{} l l @{}}
	$N \in \Rnum$ & NMPC horizon.    \\
	$N_\mathrm{g} \in \Rnum$ & Reference horizon.    \\
	$\bm{\delta} \in \Rnum^{4 \times N_\mathrm{g}} $ & Sequence of gait status.\\
	$\vect{p}_\mathrm{f} \in \Rnum^{12 \times N_\mathrm{g}}$ 
	& Footholds sequence.\\
	$\vect{x}_\mathrm{c}^\mathrm{g} \in \Rnum^{4 \times (N_\mathrm{g}+1)}$ 
	& State of the optimized reference \\ 
	& generator. \\
	$\vect{p}_\mathrm{c,k}^\mathrm{g} \in \Rnum^{2}$ 
	& X-Y COM reference position at time k. \\
	$\vect{v}_\mathrm{c,k}^\mathrm{g} \in \Rnum^{2}$ 
	& X-Y COM reference velocity at time k. \\
	$\vect{w}_\mathrm{k}^\mathrm{g} \in \Rnum^{2}$ 
	& ZMP position at time k.\\
	$\vect{s}_\mathrm{k} \in \Rnum^{2 \times 1}$
	& slack variables at time k. \\
	$\vect{u}^\mathrm{qp} \in \Rnum^{3 |\mathbb{C}|}$
	& GRFs computed by the QP Mapping. \\
	$\vect{u}^\mathrm{des} \in \Rnum^{n_u \times N}$ 
	& Predicted GRFs by the NMPC.\\
	$\vect{x}^\mathrm{des}_\mathrm{c} \in \Rnum^{12 \times (N+1)}$ 
	& Predicted states by the NMPC.\\
	$\vect{x}^\mathrm{act}_\mathrm{c} \in \Rnum^{12}$ 
	& Actual robot state.\\
	$\bar{\vect{p}}_\mathrm{c} \in \Rnum^{2}$ 
	& Average X-Y COM position.
\end{tabular}
\section{Introduction}\label{sec:introduction}
\vspace{2mm}
\subsection{Related Work}
Legged robots are becoming popular nowadays, thanks to their ability to operate on irregular and complex terrains. 
The challenge is represented by the design of a proper control strategy that allows the robots to execute their tasks.
Early developed methods involve the use of heuristic approaches, e.g.,~\cite{raibert1986legged}.
They demonstrated good performance and succeeded in hardware experiments, but are tailored to specific motions and scenarios. 
More recently, Trajectory Optimization (TO) techniques~\cite{Winkler2018a, Bratta2020, Li2020} were introduced, 
since constraints and cost functions can ensure dynamic feasibility and desired performance.
In particular, \gls{mpc} approaches can compensate for uncertainties and changes in the environment, by
computing a new trajectory online, while the robot moves. 
In our previous work~\cite{openaccess}, we presented an \gls{nmpc} formulation, which runs at 25~$\mathrm{Hz}$ and allows the 
\gls{hyq} robot~\cite{semini11hyqdesignjsce} to perform omni-directional motions, detect a pallet and step on it with improved leg mobility.
Minniti et al.~\cite{minniti22} integrated Control Lyapunov Functions into their \gls{mpc} to guarantee stability when the robot has to interact with unknown objects. 
Hong et al.~\cite{Hong} presented an \gls{nmpc} implementation with a set of different gaits, while Amatucci et al.~\cite{Amatucci2022}  
exploited a Monte Carlo Tree Search to optimize also for the contact schedule.
The last two works, however, have only been tested in simulation.\\
\begin{figure}[!t]
	\centering
	\includegraphics[scale=0.116]{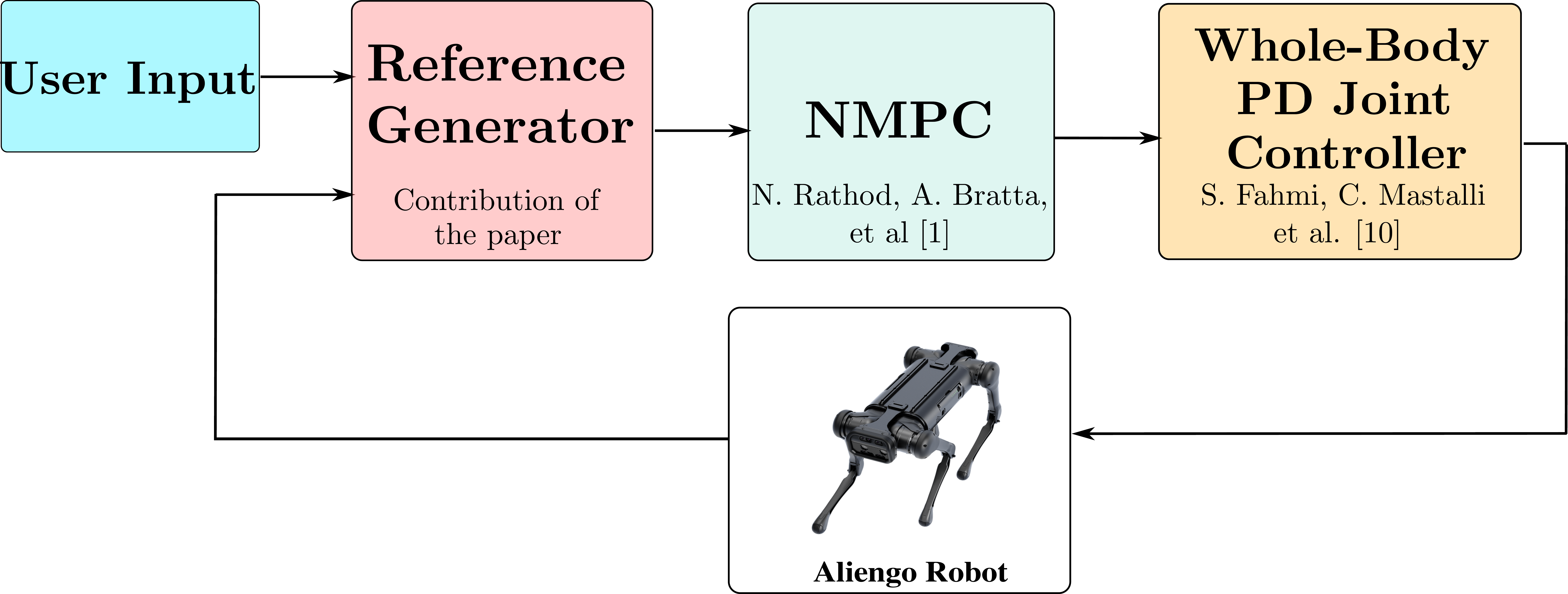}
	\caption{Overview of the locomotion framework. The Reference Generator presented in this work is integrated with the \gls{nmpc} presented in our previous work~\cite{openaccess}.
		The controller introduced in~\cite{Fahmi2019RAL} allows the robot to track the trajectories computed by the \gls{nmpc}. We used the quadruped robot Aliengo for the simulations and experiments.}
	\label{fig:aliengo}
\end{figure}
The importance of the \gls{mpc} approaches is well evident in the case of disturbances,
since they 
allow the robot to comply with external pushes (e.g., in ~\cite{Meduri2022})
thanks to high-frequency re-planning.
One of the shortcomings of a standard \gls{mpc} implementation is related 
to the fact that the robot becomes ``transparent'' to external pushes,
i.e., at each sample the new trajectory starts from the \textit{actual} position.
Therefore an additional effort is required
to track a specific \textit{reference}
Cartesian position or to return, after a push, to the original one.
A common practice is to have user-defined 
values as a reference for the \gls{mpc}~\cite{DiCarlo2018}, but in these approaches 
the user would have to manually change the reference velocity in order to compensate
for the deviation caused by the push.\\
Another possible  solution would be to add in parallel to the \gls{mpc} a simple Cartesian \gls{pd} control action 
to attract the \gls{com} to the reference position~\cite{Grimminger2020}.
However, this control strategy has some notable limitations, in particular: 
1) the added wrench does not consider the wrench produced by the \gls{mpc} 
layer (hence, there will be a fight between the two controllers with the risk of violating feasibility), 
2) the \gls{pd} is not aware of the hybrid dynamics of the legged robot, i.e.,
the intermittent contacts and the (possible) under-actuation.
Moreover, footholds play an important role in the robot agility,
since their coherence with the \gls{com} can improve the stability when the robots 
has to ``react" to an external disturbance.
3) Finally, another drawback of \gls{pd} approaches is that they cannot change feet trajectories, 
which is crucial to deal with lateral pushes.  
For example, %
Barasuol et al.~\cite{Barasuol2013} presented 
a \textit{Push Recovery} module which modifies the footholds to counteract the disturbance 
and track the position with the \gls{pd}. As explained, the drawback of this approach is that the computed locations of the feet 
give no guarantee that the resulting wrench can be generated by the robot.\\
Another possibility to track a fixed goal is adding it in the cost function. 
This would solve the issue number 1 of the \gls{pd} controller, 
because the constant position is embedded in the cost function of the \gls{mpc}, hence the ``conflict'' is dealt with at the cost level 
resulting in feasible contact forces.
However, this solution suffers from the fact that it is 
only able to apply a limited resistive force before the 
legs lose their control authority (e.g., when the \gls{com}/\gls{zmp} goes out of the support polygon), 
and therefore is of restricted applicability.
Also in this case, footholds are not generally designed to be consistent with the resulting \gls{com} trajectory, since they are computed with simple heuristics. 
Cebe et al.~\cite{Cebe2020} optimize for \gls{com} and footholds, but they can re-plan only at the touchdown moments, due to the high computational time required.\\
In addition, the \gls{mpc}s usually employ references for contact forces that come from crude 
heuristic computations (i.e., dividing the gravity weight along the legs based on static conditions). 
These values (i.e., purely vertical forces) are often not feasible 
for the
motion of the base. 
Undoubtedly, accurately tuning the weights for the different cost terms is a tedious task~\cite{Bouyarmane18}
and having more 
physically meaningful references has been shown to be a preferable solution~\cite{Bledt2019a}.
To address this problem, Bjelonic et al.~\cite{Bjelonic2022} use the result of an offline TO as cost terms for their \gls{mpc}.\\
In this work, we propose a novel optimization-based reference generator layer that supplies a \gls{nmpc} with references for 
\gls{com} and \gls{grfs} that are suitable to the task of the robot.
The novelty of our approach is that the reference trajectories are computed \textit{online} (differently from~\cite{Bjelonic2022}), 
solving a simplified optimization problem that takes into account the future robot behaviour, and the intermittent 
contact schedule to generate the references.
It is worth highlighting the difference of our approach with the Reference Governors
~\cite{Bemporad98, Kolmanovsky14, Garone17}. In fact, a Governor filters the set-points 
(i.e., it takes a reference signal vector and changes it to another reference signal of the same type) just to enforce state and input constraints.
In the proposed approach, instead,
we are interested in considering the intrinsic \textit{under-actuation} of a trot
(chosen as a template gait)
when only two feet are in contact with the ground.
Moreover, we impose additional features, such as a desired interval 
in which the robot has to reach a fixed goal or recover its position automatically from big pushes.
Another advantage of our algorithm is that it is able to affect \gls{com} trajectories and footholds at the same time, 
even though the latter are not directly included among the optimization variables.\\
A possible drawback of such a cascade optimization setting is represented by the computational effort, 
which can result in a reduction of the \gls{nmpc} frequency.
Aiming to find a compromise between accuracy and computational efficiency, 
we employ models of different complexity using the simplest \gls{lip}
~\cite{Kajita01} in the reference generator and a 
more complex one (the \gls{srbd}~\cite{Orin2013a}) for the \gls{nmpc}.
The latter, in fact, computes the state trajectories and \gls{grfs} 
that are then sent to the robot, so it requires more accuracy.
Thanks to the optimal references, 
we can avoid using a full dynamics model as in~\cite{Mastalli2022}.
\subsection{Proposed Approach and Contribution}
In this work we introduce our optimization-based reference generator
which endows the cost function of an \gls{nmpc}~\cite{openaccess}  with physically 
informed reference trajectories to be tracked.
We integrated this module with our previous works, see Fig \ref{fig:aliengo}.\\
To summarize, the contributions of the paper are:
\begin{itemize}
	\item the presentation of a novel reference generator which
	drives the robot to accomplish a task 
	(optionally in a user-defined time interval), 
	taking into account the under-actuation 
	of statically unstable gaits, like the trot.   
	Footholds are heuristically computed to be coherent 
	with the \gls{com} motion, and optimized \gls{grfs} are obtained in order to follow
	those trajectories. 
	The formulation is lightweight enough to maintain the re-planning frequency of  25 $\mathrm{Hz}$ of the \gls{nmpc}.
	\item simulations and experiments to demonstrate the effectiveness of the proposed approach in three different scenarios: 
	(a) straight motion, (b) fixed lateral goal and (c) recovery after a push. 
	We also compared in simulation our algorithm with a state-of-the-art approach (\gls{nmpc} + PD action)
	for the scenario (c).
	\item as an additional minor contribution, we demonstrate the generality of the approach 
	showing it was able to deal with different dynamic gaits, i.e., trot and pace.
\end{itemize}
\subsection{Outline}
The paper is organized as follows: Section \ref{sec:locomotion} gives an
overview of our planning framework, highlighting the main features of the reference generator.
Section \ref{sec:opt_ref_gen} describes the optimization problem with the \gls{lip} 
model and how it is used to compute \gls{com} position velocity and \gls{grfs} references.
Simulations and experiments with our Aliengo\footnote{https://www.unitree.com/products/aliengo/}
robot are presented in Section \ref{sec:results}.
Finally, we draw the conclusions in Section \ref{sec:conclusion}.
\section{Locomotion Framework Description}
\label{sec:locomotion}
\begin{figure*}[!t]
	\centering
	\includegraphics[width= \textwidth]{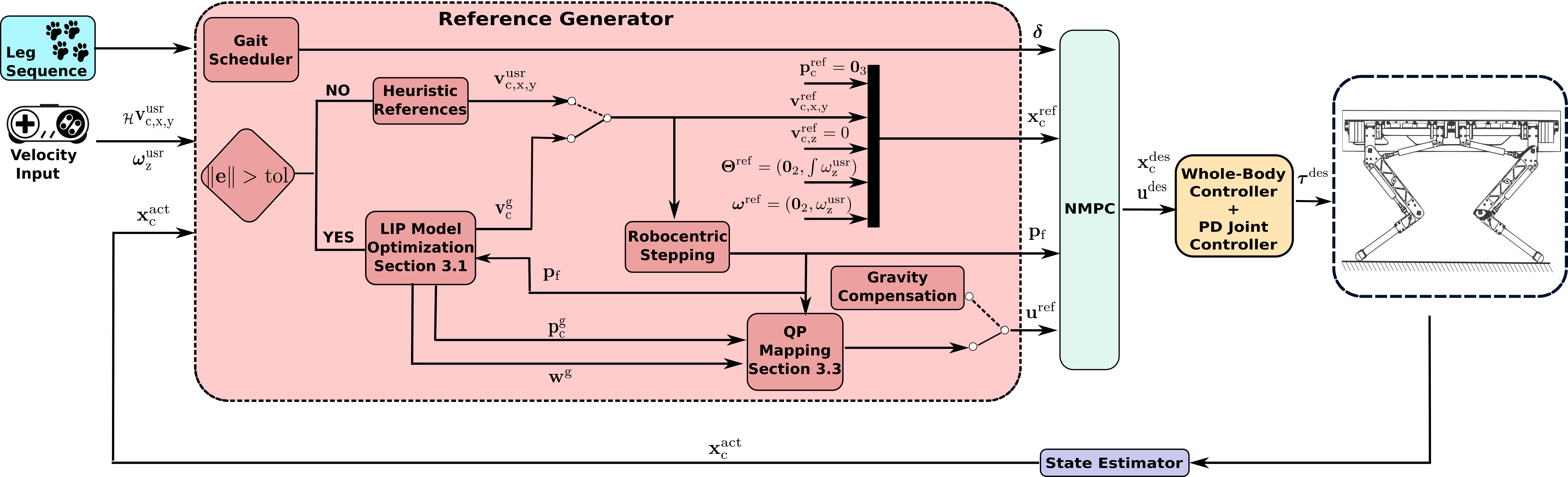}
	\caption{Block diagram of the proposed locomotion framework. Given the leg sequence, the user-defined velocities, and the actual state of the robot $\vect{x}^\mathrm{act}_\mathrm{c}$, 
		the reference generator computes the gait status $\bm{\delta}$, the footholds  $\vect{p}_\mathrm{f}$, the reference states 
		$\vect{x}^\mathrm{ref}_\mathrm{c}$ and \gls{grfs} $\vect{u}^\mathrm{ref}$ for the \gls{nmpc} \cite{openaccess} at each control loop.
		If the norm of the error $\vect{e}$  between the goal and the average \gls{com} position during the last gait cycle
		$\bar{\vect{p}}_\mathrm{c}$
		is smaller than the threshold, simple heuristic techniques are used (see \cite{openaccess}).
		In the other case, the reference generator is in the \textit{optimize} status.
		\gls{com} position and velocity to accomplish the task are computed with a \gls{lip} model; a QP mapping is used 
		to obtain the \gls{grfs} corresponding to the ZMP computed by the \gls{lip} optimization. 
		In addition, the \gls{com} velocity is used to improve the footholds.
		Finally, the \gls{nmpc} computes the \textit{optimal} trajectories for \gls{com} quantities $\vect{x}^\mathrm{des}_\mathrm{c}$ and \gls{grfs} $\vect{u}^\mathrm{des}$. 
		A Whole-Body Controller and a PD-Joint controller compute the optimal torques $\bm{\tau}^
		\mathrm{des}$ that are sent to the robot.}
	\label{fig:scheme}	
\end{figure*} 
Figure \ref{fig:scheme} gives an overview of our entire locomotion framework. 
The user decides the leg sequence and the values of both linear $\vect{v}_\mathrm{c}^\mathrm{usr} 
\in \Rnum^2$ and heading $\omega^\mathrm{usr}_\mathrm{z} \in \Rnum$ \gls{com} reference velocities, i.e., the velocities that the robot should follow. 
The velocities can be changed during the motion and the \gls{nmpc} will immediately react accordingly. 
In this work, we use two modules already presented in~\cite{openaccess}: 
the \textit{gait scheduler} and the \textit{robocentric stepping}
(also used in~\cite{Focchi2020}).\\
Given the user-defined leg sequence, 
the gait scheduler returns the gait status (either swing or stance) 
$\bm{\delta}_{i,k} 
\in
\Rnum$ of each leg $\mathrm{i}$ and for each time instant $\mathrm{k}$ 
in the \textit{reference} horizon $N_\mathrm{g}$,
re-conciliating it with the real condition of the robot, e.g., early or late touchdown. 
The reference horizon $N_\mathrm{g}$ corresponds to the maximum response time 
$T_\mathrm{f}$ of the reference generator. 
Note that $N_\mathrm{g}$ 
can be different from the horizon $N$ used in the \gls{nmpc}, with $N_\mathrm{g} \geq N$.
We use the symbol $\bm{\delta} \in \Rnum^{4 \times N_\mathrm{g}}$ to refer to 
the entire sequence of gait status.
The robocentric stepping module, on the other hand, has the 
important task of computing footholds
for each leg over the entire horizon. 
We use the symbol  
$\vect{p}_\mathrm{f} 
\in \Rnum^{12 \times N_\mathrm{g}}$ to denote these quantities for the whole horizon.\\
The key idea of robocentric stepping is that the touchdown points are computed with respect to the hip position 
and they are offset, with respect to that, depending on the \gls{com} reference velocity. 
Applying the robocentric stepping with the optimized reference velocities
allows us to obtain the coherence between \gls{com} and footholds.
Variables $\vect{p}_\mathrm{f}$ and $\bm{\delta}$ are also used as parameters in the \gls{srbd} model of the \gls{nmpc}.
If the error between the goal and average \gls{com} position is lower than a threshold (see Section \ref{subsec:goal}), 
the reference generator does not perform optimizations and user-defined velocities (heuristic references) are used as references also for the \gls{nmpc}.
We refer to this condition as \textit{heuristic} reference generator. \\
Instead, when the reference generator is in the \textit{optimize} state,
the sequences $\vect{p}_\mathrm{f}$ and $\bm{\delta}$ are the input parameters 
to a first-stage optimization that employs the \gls{lip} model (Section \ref{sec:lip_model}).
Here we compute the optimal X-Y \gls{com} trajectory $\vect{p}^\mathrm{g}_\mathrm{c}, \vect{v}^\mathrm{g}_\mathrm{c}$ to reach the goal 
$\vect{p}_\mathrm{c}^\mathrm{goal}$ where dynamic stability is satisfied (i.e., \gls{zmp} always inside the support polygon).
However, since the footholds are computed the first time with the simple heuristic approach, we need to 
recompute them using the optimized velocity. 
This will result in a new set of footholds that will be used as inputs for a second optimization. 
We iterate this until a stop condition is reached, 
e.g., the maximum number of iterations or difference between two consecutive solutions
below the defined threshold.
As a safety check, if, instead, the solver is not able to converge to a feasible solution, the reference generator is set back to the \textit{heuristic} status and $\vect{v}_\mathrm{c}^\mathrm{usr}$/gravity compensation are used as references. \\
The \gls{com} velocity trajectory $\vect{v}_\mathrm{c}^\mathrm{g}$ computed by the \gls{lip} model optimization is 
then used as velocity reference for the \gls{nmpc}.\footnote{as described in \cite{openaccess}, 
	reference \gls{com} position, roll and pitch are not tracked, the reference for the yaw is obtained by integrating the user-defined yaw rate $\omega^\mathrm{usr}_\mathrm{z}$.
	We use the variable $\bm{\Theta}^\mathrm{ref} \in \Rnum^3$ to indicate the references for roll, pitch and yaw.}
A \gls{qp}-based mapping (Section \ref{sec:qp}) computes the references for the \gls{grfs} $\vect{u}^\mathrm{ref} \in \Rnum^{12 \times N}$.\\
Finally, the output of the \gls{nmpc} are the \gls{com} trajectories
(position, orientation\footnote
{as explained in \cite{openaccess}, we parameterized the orientation with Euler angles hence the state dimension is 12.},
linear and angular velocities) $\vect{x}^\mathrm{des}_\mathrm{c}
\in \Rnum^{12 \times (N+1)}$ and \gls{grfs}
$\vect{u}^\mathrm{des} \in \Rnum^{12 \times N}$ 
that will be sent to the Controller block.
The Controller is composed of a 250 $\mathrm{Hz}$ \gls{wbc} \cite{Fahmi2019RAL} and a 1 $\mathrm{kHz}$ \gls{pd}-Joint controller.
They generate the torque references $\bm{\tau}^\mathrm{des}$ for the low-level joint controller of the robot.
The State Estimator module \cite{nobili_camurri2017rss} provides the actual values $\vect{x}_\mathrm{c}^\mathrm{act}$ of the robot at a frequency of 500~$\mathrm{Hz}$. 
\subsection{Goal Setting and Status of the Reference Generator}
\label{subsec:goal}
As already explained in Section \ref{sec:introduction}, the robot cannot follow a user-defined velocity in the presence of non-idealities or external disturbances,\footnote{
	the case of a fixed goal corresponds to a scenario in which $\vect{v}^\mathrm{usr}_\mathrm{c,y}=~0$.}
due to the \gls{mpc} transparency.  
For this reason, we define the goal $\vect{p}_\mathrm{c}^\mathrm{goal}$ as the position the robot would have reached 
if it had followed the user-commanded velocities 
$ \vect{v}_\mathrm{c}^{\mathrm{usr}} \in \mathbb{R}^2$. %
The goal is updated
at each iteration of the \gls{nmpc}.
It is initialized with $\vect{p}^\mathrm{goal} = 
\vect{p}_\mathrm{c, x,y}^\mathrm{act} + N_\mathrm{g} 
\vect{v}^\mathrm{usr}_{\mathrm{c}} T_\mathrm{s}$
at the beginning of an experiment and at each iteration 
it is incremented by $\vect{v}^\mathrm{usr}_{\mathrm{c}} T_\mathrm{s}$. 
Variable $T_\mathrm{s}$ is the sampling time of the reference generator 
and it is equal to 
$ 1/f_\mathrm{s}$ where $f_\mathrm{s}$ is the planning loop frequency.\footnote{For the sake of simplicity, we assume that the sampling time $T_\mathrm{s}$ of the \gls{nmpc} and the reference generator are the same.} 
As an alternative, the user can decide a fixed goal, either for one coordinate or both. 
A SLAM algorithm~\cite{Durrant06, Belter16} goes beyond the scope of this paper, so we assume that there are no obstacles and the goal can be reached.\\
In order to change the status of the reference generator from 
\textit{heuristic} to \textit{optimize}
in a meaningful way (e.g., no transition with the normal sway of the robot) 
we consider the \textit{average} $\bar{\vect{p}}_\mathrm{c}$ of the X-Y \gls{com} position during the last gait cycle. 
To the average position, 
we add the offset due to the desired motion in the horizon $N_\mathrm{g}$ and compare it with the goal. 
We compute the error as: $
\vect{e} = \vect{p}_\mathrm{c}^\mathrm{goal} - 
\left( \bar{\vect{p}}_\mathrm{c} + N_\mathrm{g} 
\vect{v}^\mathrm{usr}_{\mathrm{c}} T_\mathrm{s}
\right)
$ and when its norm $\Vert \vect{e} \Vert $
\footnote{Using Euclidean distance is a standard approach, but any other options, e.g., L1 or L$_\infty$ could have been valid alternatives.}
goes beyond a threshold the reference generator status is set to \textit{optimize}.\\
Algorithm 1 illustrates the pseudo-code of the different computation phases of the
reference generator.
\begin{algorithm}
	\caption{Reference Generator}\label{euclid}
	\begin{algorithmic}[1]
		\State Reference Generator status $\gets$ HEURISTIC
		\vspace{1mm}
		\State $ \vect{e} \gets \vect{p}_\mathrm{c}^\mathrm{goal} - \bar{\vect{p}}_\mathrm{c}
		- \left (
		\int_0^{N_\mathrm{g} T_\mathrm{s}} 
		\vect{v}^\mathrm{usr}_{\mathrm{c}} dt \right) $
		\If {$\Vert \vect{e} \Vert > tol $}
		\State Reference Generator status $\gets$ OPTIMIZE
		\EndIf	
		\State $\vect{p}_\mathrm{f},\bm{\delta} \gets \textit{Heuristic References}$
		\If {OPTIMIZE Reference Generator}
		\While{stop condition is not reached}
		\State $\vect{v}^\mathrm{g} \gets \textit{LIP Model Optimization}
		(\vect{p}_\mathrm{c}^\mathrm{goal}, \vect{x}_\mathrm{c}^\mathrm{act})$
		\State $\vect{p}_\mathrm{f} \gets \textit{Robocentric Stepping} (\vect{v}_\mathrm{c}^\mathrm{g})$
		\EndWhile
		\If {\textit{LIP Model Optimization} found feasible solutions }
		\State $\vect{v}^\mathrm{ref} \gets \vect{v}^\mathrm{g}$
		\State $\vect{u}^\mathrm{ref} \gets \textit{QP Mapping}
		(\vect{p}_\mathrm{c}^\mathrm{g}, \vect{w}^\mathrm{g}, 
		\vect{p}_\mathrm{f} ) $
		\Else
		\State $\vect{v}^\mathrm{ref} \gets \vect{v}^\mathrm{usr}$
		\State $\vect{u}^\mathrm{ref} \gets \textit{Gravity Compensation}$
		\EndIf
		\Else
		\State $\vect{v}^\mathrm{ref} \gets \vect{v}^\mathrm{usr}$
		\State $\vect{u}^\mathrm{ref} \gets \textit{Gravity Compensation}$
		\EndIf
		\State $\vect{p}_\mathrm{c}^\mathrm{goal} \gets  \vect{p}_\mathrm{c}^\mathrm{goal} + \vect{v}_\mathrm{c}^\mathrm{usr} 
		T_s$
		\State solve the \gls{nmpc} \cite{openaccess}
	\end{algorithmic}
\end{algorithm}
\subsection{Formal Guarantees on Response Time} \label{sec:response_time}
\begin{figure} 
	\centering
	\includegraphics[scale = 0.4]{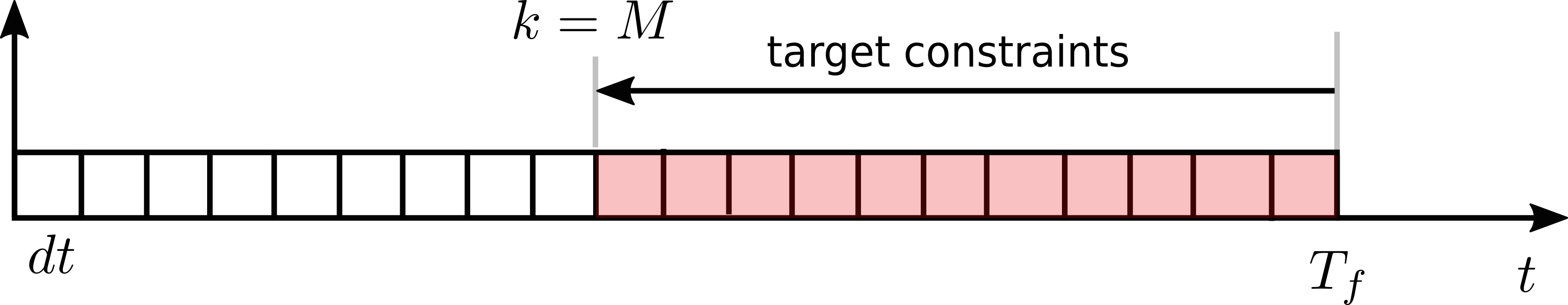}
	\caption{Pictorial representation of the goal constraints enforced to achieve a certain response time $T_\mathrm{f}$. Pink blocks corresponds to 
		the nodes of the \gls{lip} optimization in which the constraint \eqref{slack1} is active. At every iteration the variable $\mathrm{M}$
		is decreased such that the time to reach the target matches the predefined time $T_\mathrm{f}$.}
	\label{fig:shrinking_constraints}
\end{figure}

In an optimization problem, if we set the tracking of the goal as either a running cost
or a terminal cost, the response will depend on the tuning of the weights of the cost itself. 
In addition, with this approach, we cannot impose a predefined time $T_\mathrm{f}$ 
in which the robot reaches the goal (\textit{response interval}).\footnote{
	The size of the horizon $N_\mathrm{g}$ will be linked to the response time $T_\mathrm{f}$, therefore there 
	will be a maximum value of $T_\mathrm{f}$ that can be achieved by the reference generator,
	related to
	the real-time constraint posed by the re-planning frequency $f_\mathrm{s}$. 
	This will depend on the computational power of the machine where the optimization is run.}
For these two reasons a hard constraint to impose that the \gls{com} position
matches the goal can be added
to guarantee a response interval equal to $T_\mathrm{f}$, 
starting from the first time instant when the reference generator is set to the 
\textit{optimize} status. 
We also need to impose the same hard constraint on all the samples after $\mathrm{M} = T_\mathrm{f}/T_\mathrm{s}$.
Indeed, at every iteration of the \gls{nmpc}, the interval without constraint should shrink because otherwise the target would never be reached, see Fig. \ref{fig:shrinking_constraints}.
For this reason we enforce hard constraints on the goal, with the number of samples $\mathrm{M}$ that gradually reduces to zero.
In this way, we ensure to reach the goal in a time that does not depend on the tuning of the weights.
In practice, using hard constraints might lead the optimization to be trapped in an infeasible solution \cite{Hult19}.
Therefore, an implementation with slacks variables in the constraints, that allows us to relax them, is preferable.
Penalizing their value in the cost function of problem
\eqref{eq_lip}, we have the guarantee that the robot reaches the target in a 
predefined time interval and, 
if this is not possible, the solver will find the best feasible solution. 

\section{Optimized Reference Generator}
\label{sec:opt_ref_gen}
As already mentioned, the task of the reference generator is to compute along a horizon $N_g$ the linear (i.e., longitudinal and lateral) \gls{com} velocity 
reference trajectories $\vect{v}_\mathrm{c}^\mathrm{g}$ to reach a desired goal/recover from an external disturbance in a predefined 
time $T_\mathrm{f}$ and the trajectory of \gls{grfs} $\vect{u}^\mathrm{ref}$ to follow it.%

\subsection{LIP Model Optimization}
\label{sec:lip_model}
One of the features of the proposed reference generator is to take into account the intrinsic under-actuated nature of a robot when only two feet are on the ground~\cite{Chignoli20}.  
Therefore, we employ the simplest model that is able to capture this under-actuation: the \gls{lip} model \cite{Kajita01}. 
Indeed, this allows to compute the optimal trajectory for the \gls{com}, 
while imposing a desired behavior for the \gls{zmp}.
The \gls{zmp} is defined as 
``a point on the ground at which the tangential component of the moment generated by the ground reaction
force/moment becomes zero'' \cite{Harada2003}, 
and its position determines the direction and magnitude of the \gls{com} acceleration. 
Guaranteeing that the \gls{grfs} are such that the resulting \gls{zmp} 
is inside the support polygon ensures that they also satisfy 
the unilateral constraints \cite{Vukobratovic04} (the legs can only push and not pull the ground);
therefore they can be effectively realized by a real robot.
Since the support polygon boils down to a line
connecting the stance feet during a two-leg-stance phase, 
the \gls{zmp} will be able to move only on that segment.
Even though the \gls{zmp}-based models have been efficiently used in \gls{mpc} approaches, e.g., \cite{Bellicoso18}, 
we are aware that the \gls{lip} model presents some assumptions (vertical and angular dynamics are neglected).
As already mentioned, our \gls{nmpc} uses the \gls{srbd} model that will consider these dynamics in the lower optimization stage.\\
We define the state of the reference generator 
$
\vect{x}_\mathrm{c}^\mathrm{g} =
\{\vect{x}_\mathrm{c,0}^\mathrm{g}, \ldots 
\vect{x}_\mathrm{c,N\mathrm{g}}^\mathrm{g} \} \in \Rnum^{4 \times (N_\mathrm{g} +1 )}$, with 
$\vect{x}_\mathrm{c,k}^\mathrm{g} = 
\mat{ \vect{p}_\mathrm{c,k}^\mathrm{g} , \vect{v}_\mathrm{c,k}^\mathrm{g} }^T  \in \Rnum^{4}$
the stack of X-Y \gls{com} position and velocity at time $\mathrm{k}$.
The control inputs are $\vect{w}^\mathrm{g} \in \Rnum^{2 \times N_\mathrm{g}}= \{\vect{w}^\mathrm{g}_0, \ldots 
\vect{w}^\mathrm{g}_{N_\mathrm{g}-1} \} $, with 
$\vect{w}^\mathrm{g}_k \in \Rnum^{2} $ X-Y position of the \gls{zmp} at time $\mathrm{k}$. 
Footholds $\vect{p}_\mathrm{f,k}$ and gait status $\boldsymbol{\delta}_\mathrm{k}$ for all the feet at each sample $\mathrm{k}$ are the parameters 
used to compute the support polygon at each node of the reference horizon $N_\mathrm{g}$. 
Additional parameters are initial Z \gls{com} position $\vect{p}_\mathrm{c,z}^\mathrm{act}$ and $\mathrm{g} \in \Rnum = 9.81 \mathrm{m/s^2} $.
To obtain  $\vect{x}_\mathrm{c}^\mathrm{g}$ and
$\vect{w}^\mathrm{g}$ we cast the following optimization problem:
\begin{subequations}
	\begin{align} 
	\displaystyle{\min_{\vect{x}^\mathrm{g}, \vect{w}^\mathrm{g}}}\quad &
	\sum_{k=0}^{N_\mathrm{g}}  
	\parallel \vect{p}_\mathrm{c,k}^\mathrm{g} - \vect{p}_\mathrm{c}^\mathrm{goal} \parallel^2_{\mx{Q}_\mathrm{p}} +
	\parallel \vect{v}_\mathrm{c,k}^\mathrm{g}\parallel^2_{\mx{Q}_\mathrm{v} }  +
	\label{eq:cost_position_no_slip}
	\\ 
	\quad & \sum_{k=0}^{N_\mathrm{g}-1} 
	\parallel \vect{w}_\mathrm{k}^\mathrm{g} - 
	\vect{w}_\mathrm{k}^\mathrm{ref}
	\parallel^2_{\mx{Q}_\mathrm{w}} \label{eq:cost_function_no_slip}\\
	\mathrm{s.t.} 
	\quad &  \vect{x}_\mathrm{c,0}^\mathrm{g} = \vect{x}_\mathrm{c,x,y}^\mathrm{act}
	\label{init_no_slack}\\
	\quad & \vect{x}_\mathrm{c,k+1}^\mathrm{g} =
	\vect{x}_\mathrm{c,k}^\mathrm{g} + \mat{ \vect{v}_\mathrm{c,k}^\mathrm{g} T_\mathrm{s} +
		\frac{T_\mathrm{s}^2 \mathrm{g}} {2 \vect{p}_\mathrm{c,z}^\mathrm{act}}
		\left (
		\vect{p}_\mathrm{c,k}^\mathrm{g} - \vect{w}_\mathrm{k}^\mathrm{g}
		\right ) \\
		\\
		\frac{\mathrm{g}} {\vect{p}_\mathrm{c,z}^\mathrm{act}}  \left (
		\vect{p}_\mathrm{c,k}^\mathrm{g} - \vect{w}_\mathrm{k}^\mathrm{g}
		\right ) T_\mathrm{s}} 
	\label{acc_zmp_no_slack}\\
	\quad & \vect{w}_\mathrm{k}^\mathrm{g} \in 
	\mathcal{S}(\vect{p}_\mathrm{f,k},\boldsymbol{\delta}_\mathrm{k}) \quad  \quad \quad 
	k\in\mathbb{I}_0^{N_\mathrm{g}-1}
	\label{supp_polygon_no_slack}
	\end{align}
	\label{eq:lip_noslacks}
\end{subequations} 
The terms \eqref{eq:cost_position_no_slip}  and \eqref{eq:cost_function_no_slip}
of the cost function 
aims to minimize the distance between \gls{com} position and the goal,
the norm of the velocities and the distance of the \gls{zmp} 
from the center of the support polygon $\vect{w}_\mathrm{k}^\mathrm{ref}$ (for robustness purposes). 
In addition, the minimization of velocity term \eqref{eq:cost_position_no_slip} prevents having large velocities that could determine footholds outside of the workspace of the leg due to the robocentric stepping.
Matrices $\mx{Q}_\mathrm{v} \in \Rnum^{2 \times 2}$ and $\mx{Q}_\mathrm{w} \in \Rnum^{2 \times 2}$ are positive definite weighting matrices.
The initial condition \eqref{init_no_slack} is expressed by setting $\vect{x}_\mathrm{c,0}^\mathrm{g}$ equal to the corresponding 
values $\vect{x}^\mathrm{act}_\mathrm{c}$
received from the State Estimator.
Equation \eqref{acc_zmp_no_slack} corresponds to the discrete \gls{com} dynamics for the \gls{lip} model. 
Equation \eqref{supp_polygon_no_slack} imposes that the \gls{zmp} always lies inside the support polygon $ \mathcal{S}(\vect{p}_\mathrm{f,k},
\boldsymbol{\delta}_\mathrm{k})$.
The symbol $\mathbb{I}_0^{N_\mathrm{g}-1}$ indicates the set of integer numbers in the closed interval
[0 ,  $N_\mathrm{g}-1$].\\
The optimization problem \eqref{eq:lip_noslacks} does not have any guarantee on the response time. 
As already discussed in Section \ref{sec:response_time} a formulation with slack variable can be used 
to impose a predefined instant in which the robot has to reach the target:
\begin{subequations} \label{eq_lip}
	\begin{align} 
	\displaystyle{\min_{\vect{x}^\mathrm{g}, \vect{w}^\mathrm{g},\vect{s}}}\quad &
	\sum_{k=0}^{N_\mathrm{g}}  \parallel \vect{v}_\mathrm{c,k}^\mathrm{g}\parallel^2_{\mx{Q}_\mathrm{v}}  + 
	\sum_{k=0}^{N_\mathrm{g}-1} 
	\parallel \vect{w}_\mathrm{k}^\mathrm{g} - 
	\vect{w}_\mathrm{k}^\mathrm{ref}
	\parallel^2_{\mx{Q}_\mathrm{w}} +\label{eq:tracking_cost} \\
	\quad & \sum_{k=0}^{N_\mathrm{g}} \left (  
	\parallel \vect{s}_\mathrm{k}\parallel^2_{\mx{Q}_{\mathrm{s,q}}} +
	\mx{Q}_{\mathrm{s,l}} \vect{s}_\mathrm{k} \right )   
	\label{eq:slack_cost_1}\\
	\mathrm{s.t.}
	\quad &  \vect{x}_\mathrm{c,0}^\mathrm{g} = 
	\vect{x}_\mathrm{c,x,y}^\mathrm{act}
	\label{eq:initial} \\
	\quad & \vect{x}_\mathrm{c,k+1}^\mathrm{g} =
	\vect{x}_\mathrm{c,k}^\mathrm{g} + 
	\mat{ 
		\vect{v}_\mathrm{c,k}^\mathrm{g} T_\mathrm{s} +
		\frac{T_\mathrm{s}^2 \mathrm{g}} {2 \vect{p}_\mathrm{c,z}^\mathrm{act}}  
		\left (
		\vect{p}_\mathrm{c,k}^\mathrm{g} - \vect{w}_\mathrm{k}^\mathrm{g}
		\right ) \\
		\\
		\frac{\mathrm{g}} {\vect{p}_\mathrm{c,z}^\mathrm{act}} 
		\left (
		\vect{p}_\mathrm{c,k}^\mathrm{g} - \vect{w}_\mathrm{k}^\mathrm{g}
		\right )
		T_\mathrm{s}
	}  
	\label{acc_zmp}\\
	\quad & \vect{w}_\mathrm{k}^\mathrm{g} \in 
	\mathcal{S}(\vect{p}_\mathrm{f,k},\boldsymbol{\delta}_\mathrm{k}) 
	\quad \quad \quad
	\quad \quad
	\mathrm{k}\in\mathbb{I}_0^{N_\mathrm{g}-1}
	\label{eq:supp_polygon}\\
	\quad &  \mat{\vect{s}_\mathrm{k,x} \\
		\vect{s}_\mathrm{k,y}
	}
	\geq 
	\mat{
		| \vect{p}_\mathrm{c,x,k}^\mathrm{g} - 
		\vect{p}_\mathrm{c,x}^\mathrm{goal} | \\
		| \vect{p}_\mathrm{c,y,k}^\mathrm{g} - 
		\vect{p}_\mathrm{c,y}^\mathrm{goal} | 
	}
	\qquad
	\mathrm{k}\in\mathbb{I}_\mathrm{M}^{N_\mathrm{g}} %
	\label{slack1}\\
	& \vect{s}_\mathrm{k,x},  \vect{s}_\mathrm{k,y} \geq 0 \qquad\qquad 
	\qquad \quad 
	\mathrm{k}\in\mathbb{I}_0^{N_\mathrm{g}}
	\label{eq:positive_slacks}
	\end{align}
\end{subequations}
The slack variable $\vect{s}_\mathrm{k} \in \Rnum^{2 \times 1}$ is added in the cost term \eqref{eq:slack_cost_1} and thanks to Equation \eqref{slack1}, 
and \eqref{eq:positive_slacks}
it allows to impose that the robot \gls{com} coincides with the goal after $\mathrm{M}$ samples. 
Differently from the optimization problem \eqref{eq:lip_noslacks},
the cost function does not explicitly include a position term, since it is 
enforced with slacks into Equation~\eqref{eq:slack_cost_1}.
Matrices $\mx{Q}_\mathrm{s,q} \in
\Rnum^{2 \times 2}$ and $\mx{Q}_\mathrm{s,l} \in
\Rnum^{1 \times 2}$ are the additional weights for the quadratic and linear term for the slack variables. 
Equation \eqref{eq:initial}, \eqref{acc_zmp}, \eqref{eq:supp_polygon}
are the same as Equation
\eqref{init_no_slack}, \eqref{acc_zmp_no_slack}, 
\eqref{supp_polygon_no_slack} respectively.
\subsection{QP Mapping} 
\label{sec:qp}
The reference generator computes \gls{com} trajectories 
which the \gls{nmpc} must follow, but the latter also requires
reference \gls{grfs} $\vect{u}^\mathrm{ref} \in \Rnum^{12 \times N}$.
Since the output of the optimization problem \eqref{eq:lip_noslacks} or \eqref{eq_lip} is a trajectory for the \gls{zmp} $\mathbf{w}^\mathrm{g}$,
we need to map this into a set of 
consistent \gls{grfs}, 
thus moving from a bi-dimensional to the higher dimensional space of contact forces.
For this reason, we define the set of indices of the legs in contact with the ground by $\mathbb{C}$.
A parametric \gls{qp} is solved to find the vector of \gls{grfs} $\vect{u}^\mathrm{qp} \in \Rnum^{3 |\mathbb{C}|}$ 
which corresponds to the \gls{zmp} location $\vect{w}^\mathrm{g}$. 
If a foot is in the swing phase, its \gls{grfs} are set to 0 by default: 
\begin{equation} \label{eq:footposition}
\vect{u}_{\mathrm{i}}^\mathrm{ref} =
\begin{cases}
\vect{u}_{\mathrm{i}}^{\mathrm{qp}} & \mathrm{i} \in \mathbb{C} \\
\vect{0}_{3 \times 1} & \mathrm{i} \notin \mathbb{C}
\end{cases}
\end{equation}
where $\mathrm{i}$ is the leg index.
Note that we need to solve a \gls{qp} for each sample $\mathrm{k}$ in the horizon $N$, therefore 
to avoid overloading the notation, we do not specify the subscript k in the following quantities.
The parameters of the model are the footholds $\vect{p}_\mathrm{f}$, the X-Y 
components of the \gls{com} position $\vect{p}_\mathrm{c}^\mathrm{g}$,
the \gls{zmp} trajectory $\vect{w}^\mathrm{g}$,
the initial Z \gls{com} position $\vect{p}_\mathrm{c,z}^\mathrm{act}$
and the mass of the robot $\mathrm{m} \in \Rnum$. Thus,
for every $\mathrm{k}$ we solve:
\begin{subequations}\label{eq:qp}
	\begin{align}
	\displaystyle{\min_{\vect{u}^\mathrm{qp}}}\quad &
	\parallel \vect{u}^\mathrm{qp}_\mathrm{x,y}\parallel^2_{\mx{Q}_\mathrm{u}}  + 
	\left\Vert \sum_{i \in \mathbb{C}}
	\big [ \vect{p}_\mathrm{f,i} - 
	[\vect{p}_\mathrm{c}^\mathrm{g}, \vect{p}_\mathrm{c_z}^\mathrm{act}]^T
	\big ]_
	\mathrm{x}
	\vect{u}^\mathrm{qp}_\mathrm{i}\right\Vert ^2_{\mx{Q}_\mathrm{k}}
	\label{eq:cost_mapping}
	\\
	\mathrm{s.t.}
	\quad & \vect{w}_\mathrm{x,y}^\mathrm{g} = \frac {\sum_{i \in \mathbb{C}} \vect{p}_\mathrm{f_i,x,y} \vect{u}^\mathrm{qp}_\mathrm{i,z}}
	{\sum_{i \in \mathbb{C}} \vect{u}^\mathrm{qp}_\mathrm{i,z}}
	\label{eq:zmp_xy}\\
	\quad & \sum_{i \in \mathbb{C}} \vect{u}^\mathrm{qp}_\mathrm{i,z} = \mathrm{m} \mathrm{g}
	\label{eq:zmp_gravity_comp} \\
	\quad & \frac{\mathrm{g}} {\vect{p}_\mathrm{c_z}^\mathrm{act}}
	(\vect{p}_\mathrm{c}^\mathrm{g} - \vect{w}^\mathrm{g}) 
	=
	\frac{\sum_{i \in \mathbb{C}} \vect{u}_\mathrm{i, x,y}^\mathrm{qp} }
	{\mathrm{m}}
	\label{acc}
	\end{align} 
\end{subequations}
The first term of Equation \eqref{eq:cost_mapping} is the regularization term on the X-Y components of the \gls{grfs}, 
with $\mx{Q}_\mathrm{u} \in \Rnum^{2 \times 2}$ as weighting matrix, while the second one minimizes the angular momentum rate, and it is weighted by $\mx{Q}_\mathrm{k}~\in ~\Rnum^{3 \times 3}$.
Variable $\big [ \cdot \big ]_\mathrm{x}$ represents the skew-symmetric matrix associated with the cross product.
Equation \eqref{eq:zmp_xy} is the definition of \gls{zmp},
Equation \eqref{eq:zmp_gravity_comp} represents the gravity compensation. 
Equation \eqref{acc} guarantees that 
the X-Y \gls{com} acceleration computed with the \gls{lip} model in the 
reference generator (left term) coincides with the one of the \gls{srbd} used in \gls{nmpc}.
It is worth highlighting that when the robot is on two legs, the \gls{zmp} lies on a line, and so it can only move in a 1D manifold. 
During that phase, imposing the Eq. \eqref{eq:zmp_xy} for the X component is enough to guarantee that also the Y coordinate of the \gls{zmp} respects the same constraint. 
Along the horizon $N_\mathrm{g}$, each problem is independent from the others, so they can be solved in parallel. 
In this way the computation effort remains low (a couple of ms to solve a \gls{qp} problem with linear constraints) and therefore 
the new reference generator can be integrated into our high-frequency \gls{nmpc} scheme.
\section{Simulation and Experimental Results}
\label{sec:results}
The optimization-based reference generator
endows the \gls{nmpc} planner with the capability to recover from disturbances and avoid drifting in the face of non-idealities.
To show its effectiveness, we tailored three template scenarios: (a) motion along a straight line, (b) reaching a fixed goal, and (c) recover from external pushes.
We performed the simulations and experiments on the 22 $\mathrm{kg}$
quadruped robot AlienGo of Unitree.\\
We consider the trot as a template gait for our experiments  because of its inherently unstable nature.
Indeed, any asymmetry in the real robot 
can make it drift when setting a pure forward velocity. 
The quadruped is thus not able to follow a straight line.\\
The variable $T_\mathrm{s}$ is set to 40 $\mathrm{ms}$ and 
the horizons of both reference generator ($N_\mathrm{g}$)
and \gls{nmpc} ($N$) are 50 nodes,
corresponding to an interval of 2 $\mathrm{s}$.
We keep, thus, the same planning loop frequency of 
$25 \mathrm{Hz}$ as our previous work.
The trot parameters are cycle time $T_\mathrm{c}$= 1 $\mathrm{s}$,
duty factor $D$ = duration of stance phase / $T_\mathrm{c}$ = 0.65. 
The values of the weighting matrices are reported in Table 
\ref{tab:weights}. 
Their value has been tuned with a trial and error procedure.
We refer to Table 2 of~\cite{openaccess} for the values of the weights of the NMPC and the WBC + PD Joint Controller.
Without any lack of generalization, slack variables are considered only for the Y component.\\
\begin{table}[t!]
	\caption{ Weights used in the Governor}
	\begin{center}
		\begin{tabular}{@{} l l l @{}}
			\toprule[0.4mm]
			\textbf{Cost} \quad & \textbf{Weight} & \textbf{Value} \\ 
			\midrule	
			\multirow{1}{3cm}{Velocities LIP} 			
			&$\mx{Q}_\mathrm{v}$  & $\mathrm{diag}(200, 300$)   \\
			\midrule
			\multirow{1}{3cm}{\gls{zmp}}
			&$\mx{Q}_\mathrm{w}$	 & $\mathrm{diag}(100, 350)$ \\
			\midrule
			\multirow{2}{3cm}{Slack}
			&$\mx{Q}_\mathrm{s,q}$	  & $\mathrm{diag}(0, 1000)$ \\
			&$\mx{Q}_\mathrm{s,l}$ 	  & $[0, 1000]$\\
			\midrule
			\multirow{2}{3cm}{Forces QP mapping}
			&$\mx{Q}_\mathrm{u}$  & $\mathrm{diag}(100, 100)$   \\
			\\
			\midrule
			\multirow{2}{3cm}{Angular Momentum Rate}
			&$\mx{Q}_\mathrm{k}$	 & $\mathrm{diag}(1, 1, 1)$ \\
			&\\
			\bottomrule[0.4mm]
		\end{tabular}
		\label{tab:weights}
	\end{center}
\end{table}
We used the HPIPM~\cite{Frison2020} solver integrated into
$\texttt{acados}$~\cite{Verschueren2019} library to find the solutions of the problem \eqref{eq_lip}.
The problem \eqref{eq:qp}, instead, is solved using \texttt{eiquadprog} \cite{Guennebaud2011}, since it offers a more straightforward interface for the \gls{qp}s.
Both reference generator and \gls{nmpc} run on
an Intel Core i7-10750H CPU @~2.60~$\mathrm{GHz}$.
The LIP Model Optimization takes $2$-$4 \, \mathrm{ms}$ 
with the prediction horizon of $2 \, \mathrm{s}$ and thus $N$ equal to $50$ , while the time required by each QP (Eq. 4) is negligible, around 0.01 ms.\\
In this section, we will call \textit{reference} the output of the reference generator, \textit{desired} the output of the \gls{nmpc},
and \textit{actual} the real values measured by the State Estimator.
\subsection{Simulations}
\label{sec:simulation}
\begin{figure}[!t]
	\centering
	\includegraphics[scale = 0.25]{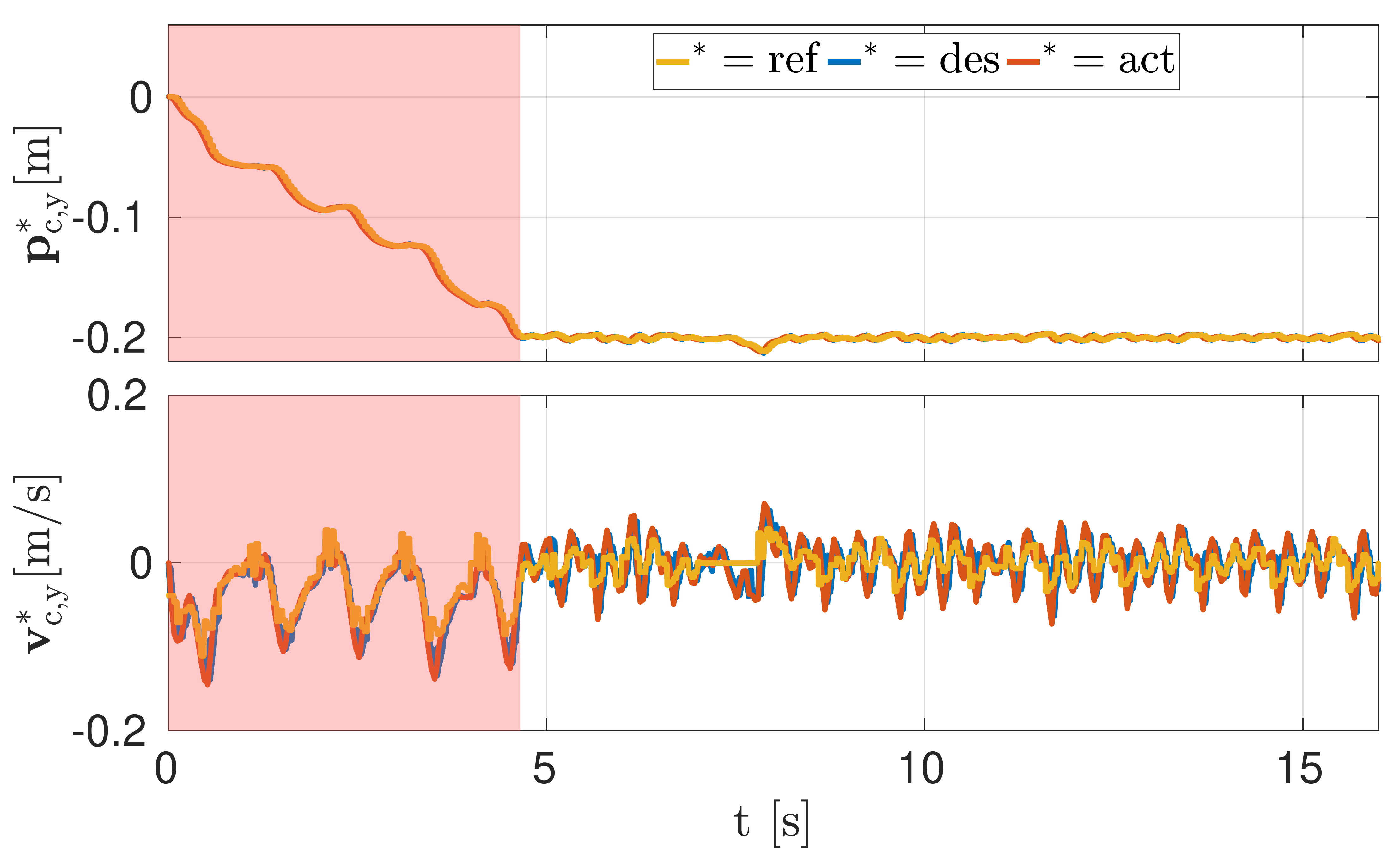}
	\caption{Simulation, scenario (b): \gls{com} Y position and velocity in a scenario in which the robot has to reach a target of -0.2 $\mathrm{m}$, 
		with a response interval of 4.8 $\mathrm{s}$ (120 iterations of the \gls{nmpc}, pink blocks).
		Yellow lines are the output of the reference generator, 
		which is used as reference for the \gls{nmpc}.
		Blue lines corresponds to the desired trajectories, tracked by the \gls{wbc}.
		The red lines report the actual values.
	}%
	\label{fig:cg_20_120}
\end{figure} 
The four-leg-stance phase in a walking trot is the only moment of the gait where the robot has full control 
authority and is able to track the reference velocities. Neglecting this fact would lead to failure or unpredictably longer response times.
In our algorithm, the reference generator already takes into account the under-actuation, enabling us to deal with this issue successfully. 
The simulations of the scenario (a) are only reported in the accompanying video. \footnote{\url{https://youtu.be/Jp0D8_AKiIY}}\\
Figure \ref{fig:cg_20_120} refers to the simulation of the scenario (b) in which the robot has to 
go to a lateral target ($-0.2$ $\mathrm{m}$ on the Y coordinate) in a predefined time ($T_\mathrm{f}= 4.8$ $\mathrm{s}$) and then keep it while walking.
Since X-Y directions are decoupled in the \gls{lip} model, the X component of the goal $\vect{p}_\mathrm{c,x}^\mathrm{goal}$ 
is updated at each iteration of the \gls{nmpc} according to Algorithm 1 and continuously tracked. 
The top plot demonstrates that the reference generator is able to compute reference \gls{com} trajectories (yellow line) that allow the actual \gls{com} values (red line)
to accomplish the task. 
Since the reference quantities satisfy the under-actuation of the gait,
reference velocities are nicely tracked by the \gls{nmpc} 
(respectively yellow and blue line, bottom part of Fig. \ref{fig:cg_20_120})
and consequently by the actual ones.
\begin{figure}[!h]
	\centering	
	\includegraphics[scale = 0.25]{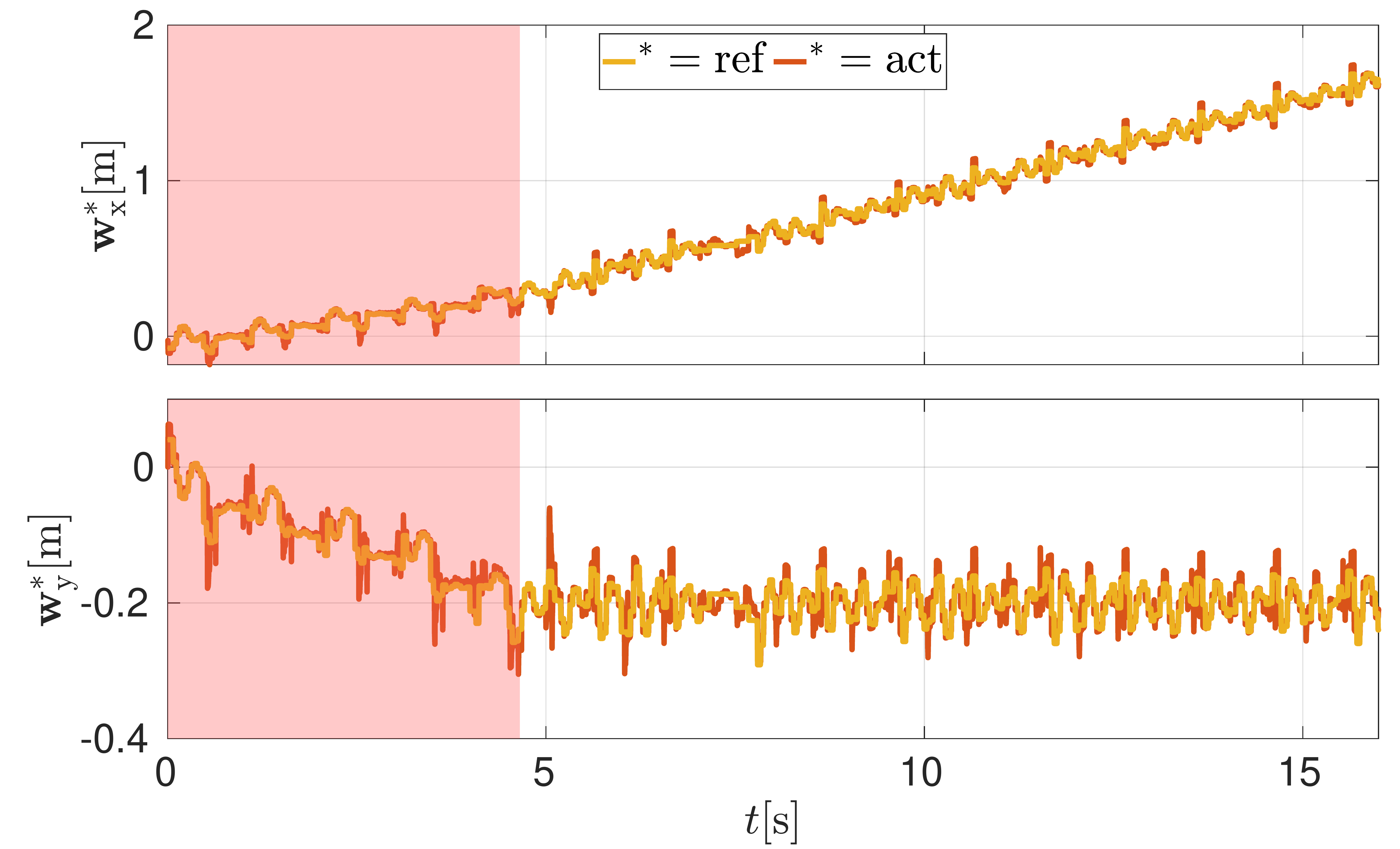}
	\caption{Simulation, scenario (b): X-Y \gls{zmp} location in a scenario in which the robot has to reach a target of -0.2 $\mathrm{m}$ along Y direction with a response time $T_\mathrm{f}= 4.8$ $\mathrm{s}$ 
		(120 iterations of the \gls{nmpc}). The yellow line corresponds to the output of the reference generator, 
		while the red has been computed using Equation \eqref{eq:zmp_xy}.}
	\label{fig:zmp_cg_20_100}
\end{figure} 
Figure \ref{fig:zmp_cg_20_100}, instead, shows the reference $\vect{w}^\mathrm{g}$ and actual X-Y components of the \gls{zmp} locations for the same simulation 
(computed by Equation \eqref{eq:zmp_xy}). 
As it can be seen, the robot is able 
to track the reference values for the entire cycle, for both X-Y coordinates, i.e., they have the same trend.
In this simulation we decided to keep the reference generator always in the
\textit{optimize} status to 
show how the algorithm is able to track (once the target has been reached) a zero user-defined lateral velocity $\vect{v}_\mathrm{c,y}^\mathrm{usr}$.\\
Figure \ref{fig:py_cg_2s} reports scenario (b) with the same goal (-0.2 $\mathrm{m}$) but different response interval ($T_\mathrm{f}= 3$ $\mathrm{s}$). %
Due to the smaller interval, the response is more aggressive and presents an overshoot which is then recovered within the 3 $\mathrm{s}$ interval.
The task is achieved by simply modifying the response interval $T_\mathrm{f}$; no tuning of the weights of the cost function is required.
Shaded areas highlight the response interval.
Figure 7 shows the tracking error between the actual and desired Y CoM position. It confirms that the error is negligible and the desired values are tracked by the robot.\\
\begin{figure}[!h]
	\centering
	\includegraphics[scale = 0.25]{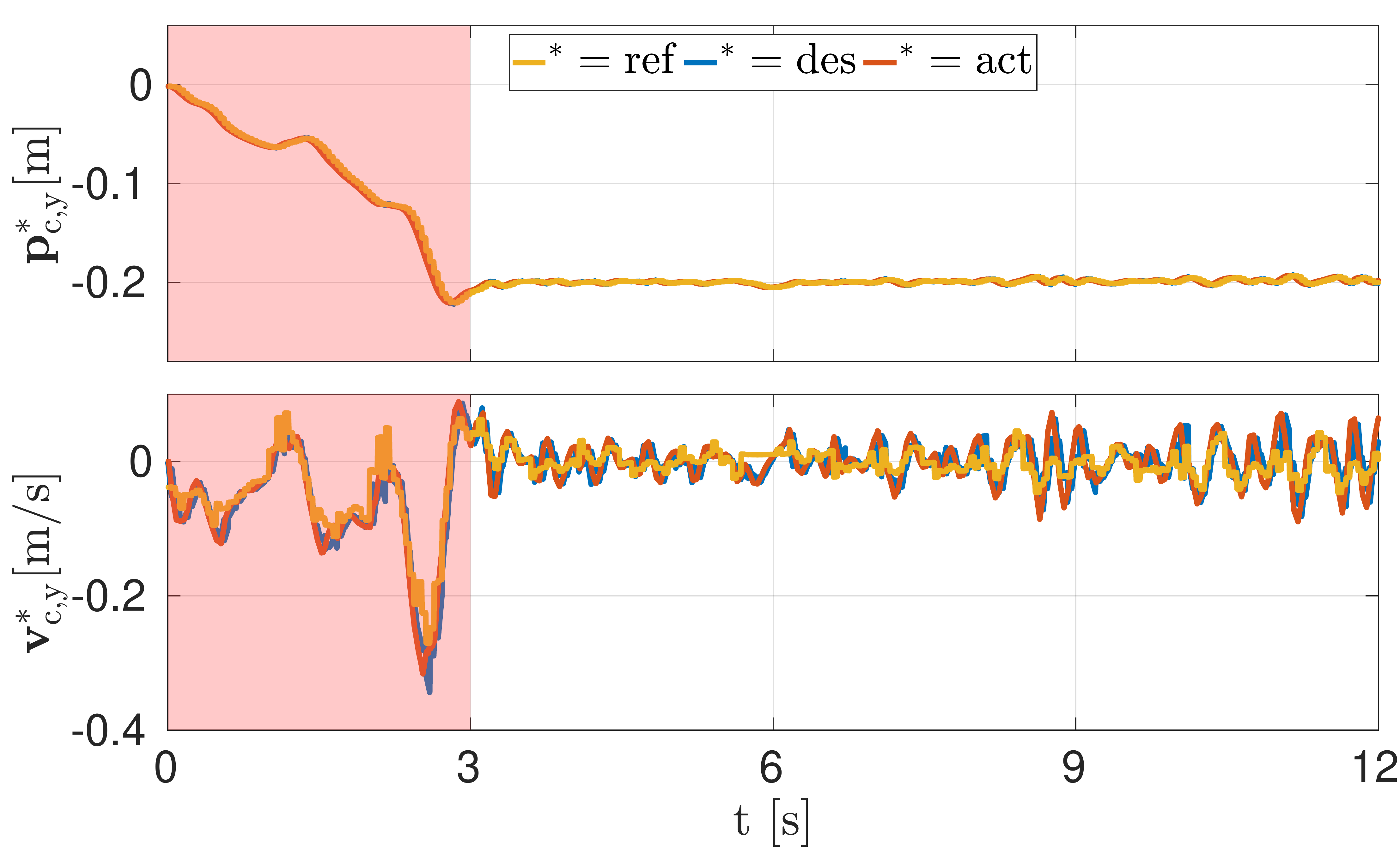}
	\caption{Simulation, scenario (b): \gls{com} Y position and velocity when the robot has to reach a target of -0.2 $\mathrm{m}$ with a response interval of $T_\mathrm{f}= 3$ $\mathrm{s}$). %
		As in Fig. \ref{fig:cg_20_120}, yellow, blue, and red lines correspond respectively to reference, desired, and actual quantities.
	}
	\label{fig:py_cg_2s}
\end{figure}
\begin{figure}[!h]
	\centering	
	\includegraphics[scale = 0.25]{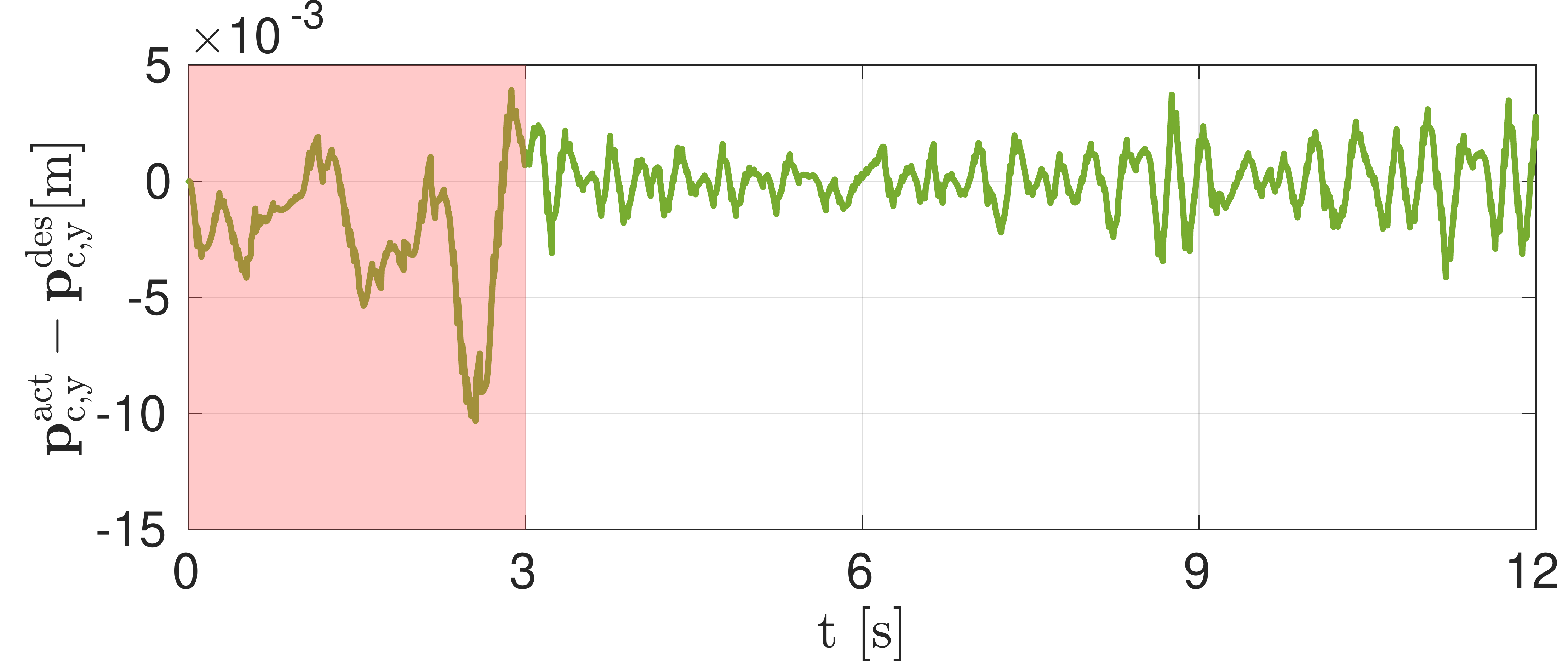}
	\caption{Simulation, scenario (b): tracking error between actual and desired Y CoM position ($\vect{p}_\mathrm{c, y}^\mathrm{act} - \vect{p}_\mathrm{c, y}^\mathrm{des}$ )} 
	\label{fig:tracking error}
\end{figure}  
Next, we want to validate our approach by comparing it with a simple cartesian 
\gls{pd} approach in the scenario (c).
This is typically implemented computing 
a feed-forward force $f_\mathrm{ff, y} \in \Rnum$ which depends on the error between the actual state and the goal, i.e., 
$f_\mathrm{ff, y} = K_\mathrm{p} 
\left (
\vect{p}_\mathrm{c,y}^\mathrm{goal} - \vect{p}_\mathrm{c,y}^\mathrm{act}
\right )
+ 
K_\mathrm{d} 
\left (
\vect{v}_\mathrm{c,y}^\mathrm{usr} - \vect{v}_\mathrm{c,y}^\mathrm{act}
\right )
$. \\
In this scenario, for the first 5 seconds of the motion the robot is pushed with a force of 15 $\mathrm{N}$ in the Y direction.
The user velocity $\vect{v}_\mathrm{c,y}^\mathrm{usr}$ is zero,
so the task is to come back to the initial Y position. 
In Fig. \ref{fig:centroidalPD} we report the actual Y \gls{com} position
$\vect{p}_\mathrm{c,y}^\mathrm{act}$ in three different cases.\\  
The green line represents the case in which the system is modeled as a second order with critically damped response.\\ 
From the control theory, we know that the settling time is equal to 
$T_\mathrm{f} = \frac {4} {\zeta \omega_\mathrm{n}}$, with $\zeta$ equal to damping ratio and 
$\omega_\mathrm{n}$ natural frequency. To have a critically damped response we aim at $\zeta$ = 1.\\
Rewriting the second order equation of the mass/spring/damper as a function of $\zeta$ and 
$\omega_\mathrm{n}$ we have
$K_\mathrm{p} = m \omega_\mathrm{n}^2$ and 
$K_\mathrm{d} = 2 \zeta \sqrt{\mathrm{m} K_\mathrm{p}}$.\\
Merging the two expressions it results $T_\mathrm{f} = 4 \sqrt{\frac{\mathrm{m}}{K_\mathrm{p}}} $ with 
$\zeta = 1$.
Considering $T_\mathrm{f} = 4.8$ $\mathrm{s}$ 
and a total robot mass $\mathrm{m} = 22$ $\mathrm{kg}$,
we obtained
$K_\mathrm{p} = 15$ $\mathrm{N}/\mathrm{m}$
and consequently
$K_\mathrm{d} = 36 \, \mathrm{Ns/m}$.\\
It is evident that the \gls{com} is far from reaching the goal in the predefined time.\\
The purple line, instead, represents the case in which the values of proportional and derivative gain are manually tuned such that the robot reaches the goal faster.
The result is $K_\mathrm{p} = 170$ $N/m$ and $K_\mathrm{d} = 122$ $\mathrm{Ns/m}$.
As it can be noticed in the plot, once the disturbance is removed,
the robot moves towards the goal, but it is not able to follow it once it has been reached.
In fact, the robot keeps drifting, moving away from the initial position. 
We can conclude thus that the approach with a feed-forward force results in a slow response, with a steady state error with respect to the goal.
The common practice of setting the \gls{pd} parameters considering the system
behaving as a second order system (i.e. spring/mass /damper) 
is not valid due to the under-actuation, showing the limitation of the approach.
In addition, even hand-tuning of the parameters does not allow the user to obtain the desired behaviour, e.g., the values of $K_\mathrm{p}$ and $K_\mathrm{d}$ which have good performance in recovering from the push (purple line) are not suitable for the straight forward motion.
The result of our approach, instead, is reported with light blue line.
Once the disturbance has been removed, the reference generator
computes a velocity trajectory that brings the robot towards the goal
in the desired time $T_\mathrm{f}$ (shaded area) and without steady state error.
\begin{figure}[!h]
	\centering
	\includegraphics[width=\columnwidth]{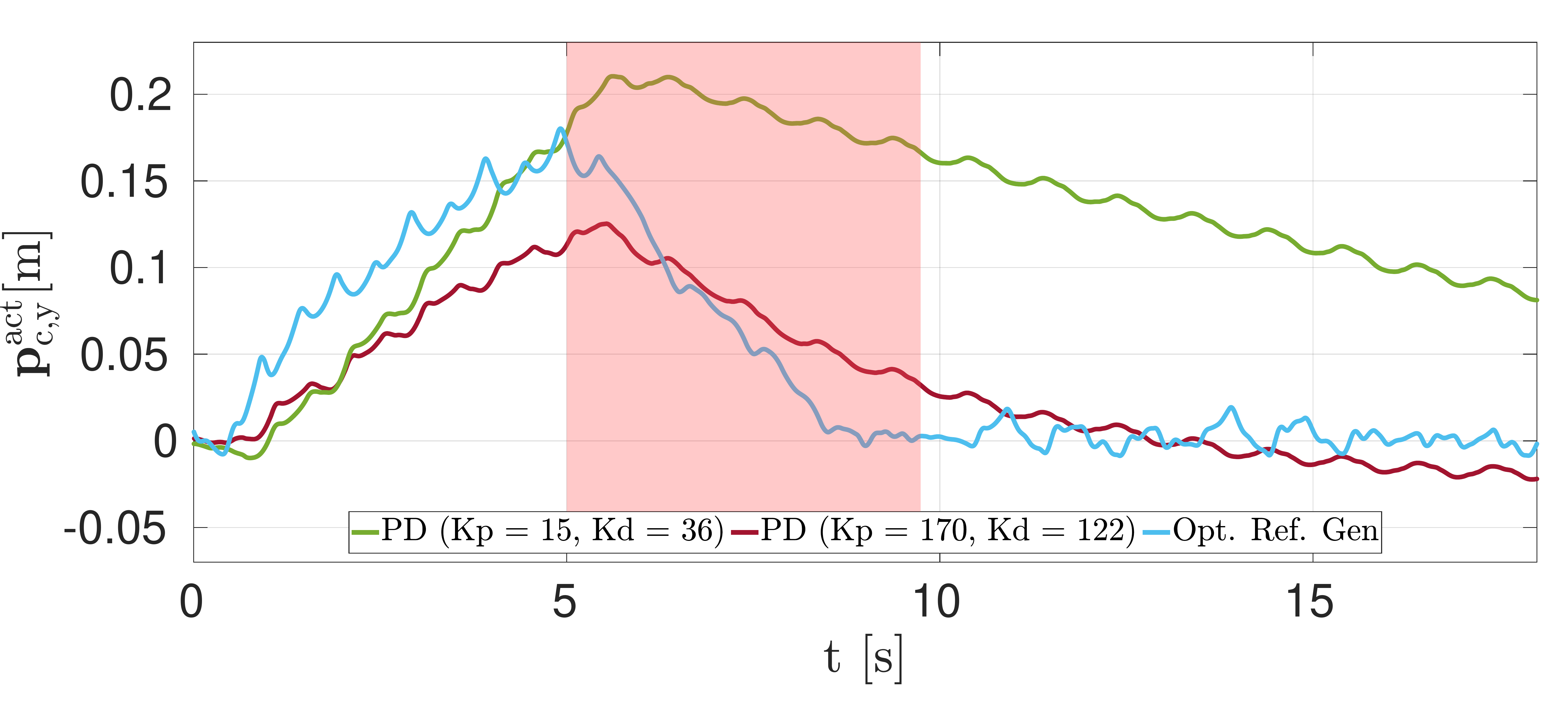}
	\caption{Simulation, scenario (c): comparison of the actual Y \gls{com} positions between the \gls{pd} force approach (green and purple lines) and our reference generator
		(light blue line). The goal is to come back to the initial position after a 15 $\mathrm{N}$ lateral push of 5 $\mathrm{s}$.
		For the green line the values are chosen to impose a response time of 4.8 $\mathrm{s}$ considering a second order system response. 
		For the purple line, instead, K$_\mathrm{p}$ and K$_\mathrm{d}$ are computed such that the system should have a critically damped response. 
		In both cases 
		the robot is not able to converge to the goal. With our reference generator (light blue line) the robot recovers its initial position after the push, without any steady-state error.
	}
	\label{fig:centroidalPD}
\end{figure}
\subsection{Experiments} \label{sec.experiments}
\begin{figure}[!h]
	\centering
	\includegraphics[scale = 0.28]{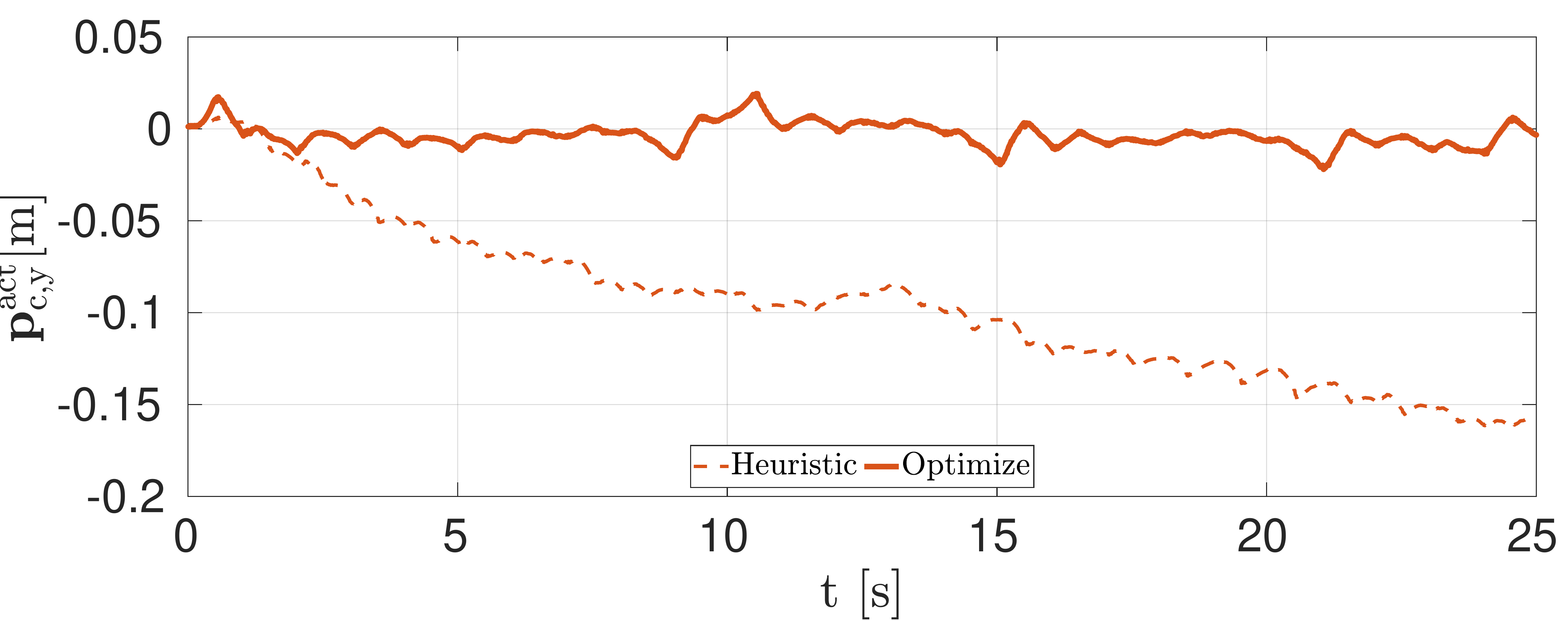}
	\caption{Experiment, scenario (a): Aliengo moving forward with zero lateral velocity set by the user. 
		The dashed line represents the actual Y \gls{com} position when the reference generator is forced to be always in \textit{heuristic} status. The robot diverges from the goal of $\vect{p}
		_\mathrm{c,y}^\mathrm{act} = 0$ and there is no part in the controller that brings it back.
		The continuous line shows the actual Y \gls{com} position  when the reference generator automatically changes 
		its state according to the error $\vect{e}$. Thanks to the corrections of the \textit{optimize} reference generator, Aliengo is able to stay close to the goal.   
	}
	\label{fig:forward_comparison}
\end{figure}
In this section we present \textit{preliminary} experiments for the scenarios (a), (b), (c) carried out with the real robot platform. 
In these experiments, we employed the problem \eqref{eq:lip_noslacks} 
without ensuring guarantees on the response time. 
As first experiment we performed scenario (a) on the robot, as illustrated in Fig. \ref{fig:forward_comparison}.
\begin{figure}[!h]
	\centering
	\includegraphics[scale = 0.3]{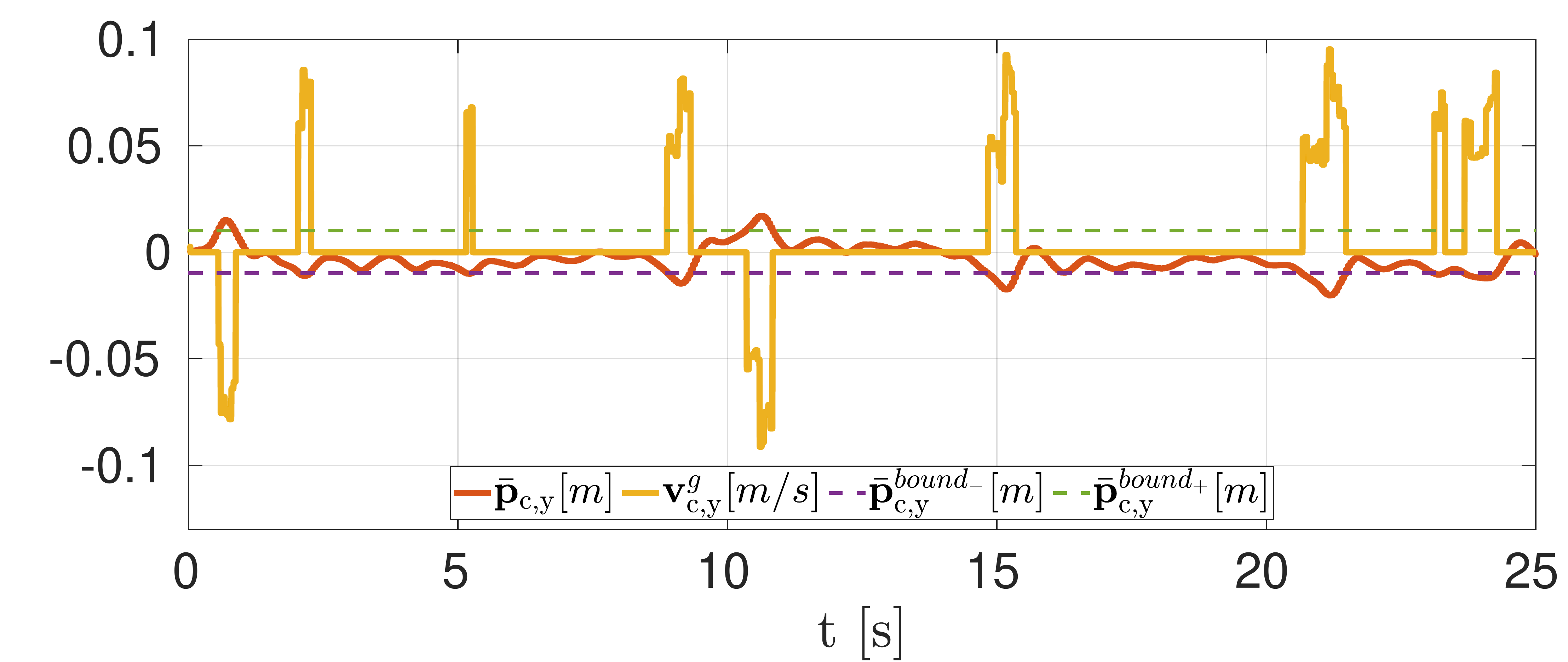}
	\caption{Experiment, scenario (a): Aliengo moving forward with zero user lateral velocity. 
		The peaks in the reference velocity (yellow line) represent the moment in which the average Y position (red line) 
		has passed the threshold (dashed lines) around the initial position and the
		reference generator is set to \textit{optimize} status.}
	\label{fig:forward}
\end{figure}
The figure demonstrates that the \gls{nmpc} with only a heuristic reference generator (dashed lines) does not succeed in moving on a straight line for a statically unstable gait as trot.

\noindent Indeed, the robot suffers lateral and backward (less visible) drifts because of two reasons: 
1) the trot being an unstable gait the \gls{com} always diverges between two four-leg-stance
events 
in opposite directions. Any little asymmetry in the robot results in a cumulative drift in one direction. 
2) since Aliengo has c-shaped legs, they create nonzero moments about the pitch axis during a swing.\\
Enabling the optimal reference generator, instead, the robot is able to reach the goal and prevent the trajectory from drifting (see the continuous line in Fig. \ref{fig:forward_comparison}).\\
\begin{figure}[h!]
	\centering
	\includegraphics[scale = 0.28]{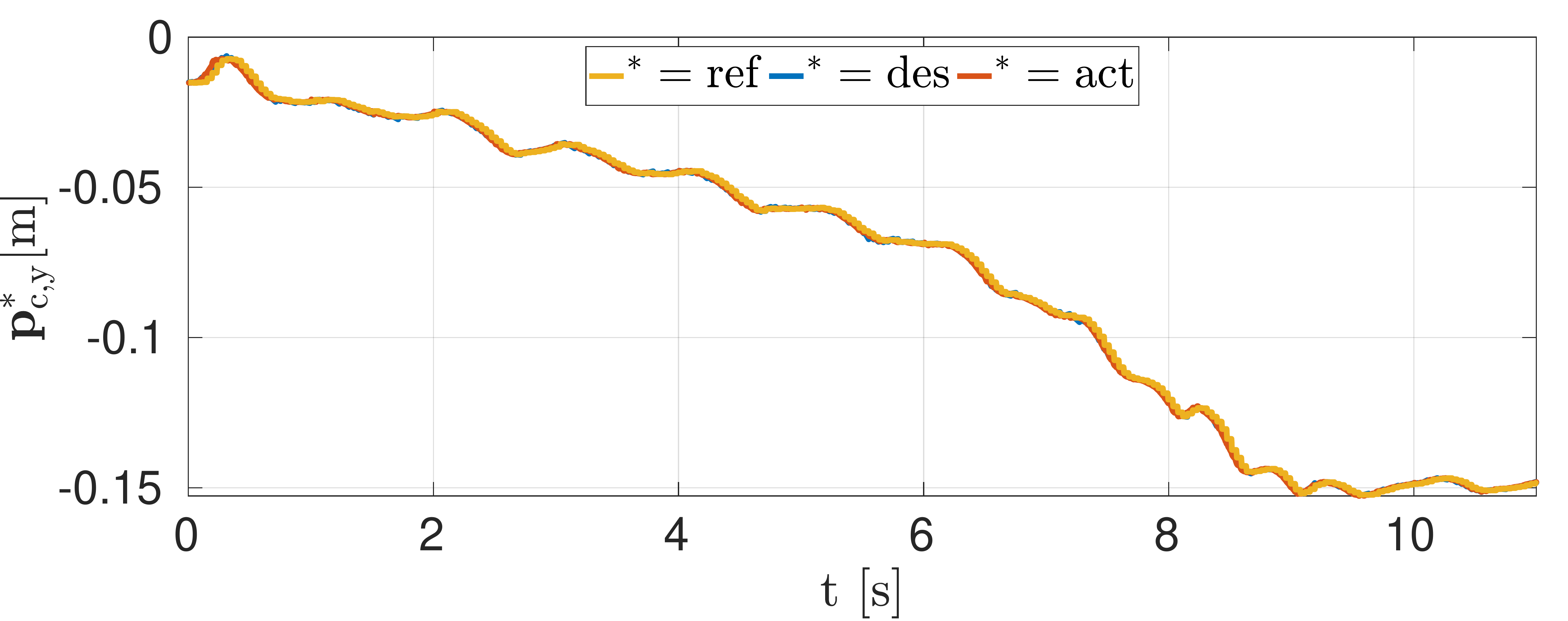}  
	\caption{Experiment, scenario (b): Aliengo robot reaches the target
		position of -0.15 $\mathrm{m}$ and then keeps moving following the user-defined velocity.}
	\label{fig:constant}
\end{figure} 
Figure \ref{fig:forward} shows the change in the reference lateral velocity (yellow line) done by the reference generator.
When the average Y position ($\bar{\vect{p}}_\mathrm{y}$, red line) exceeds the bounds ($\bar{\vect{p}}_\mathrm{y}^\mathrm{bound}$, dashed lines), 
the status is changed to \textit{optimize} and the reference generator 
brings back the \gls{com} close to the goal.
A threshold of 1 $\mathrm{cm}$ around the constant goal has been chosen for the activation. 
Once the goal has been reached, the reference generator automatically resets to \textit{heuristic}, and the reference velocity becomes equal to the user one (zero).
The continuous changing of the status of the reference generator demonstrates 
the need to have an external module 
which corrects the reference trajectories during a trot.\\ 
\begin{figure*}[!t]
	\centering
	\includegraphics[width= \textwidth]{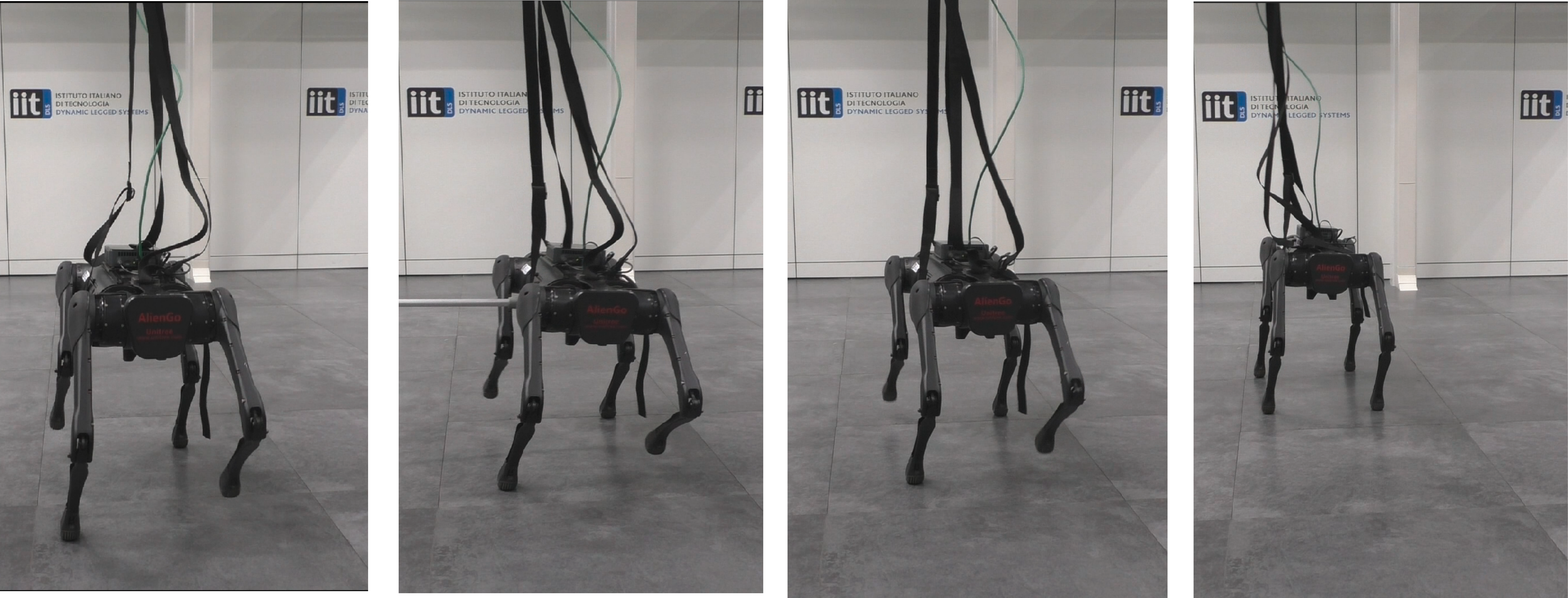}
	\caption{Experiments, scenario (c), sequence of screenshots.
		The robot moves forward (picture 1), and is suddenly pushed with a stick (picture 2). 
		Once the push is removed (picture 3), the optimized reference generator automatically drives the robot back to its initial position. 
		Finally, the robot follows the user-defined velocity (picture 4).}
	\label{fig:screenshot}	
\end{figure*}
Figure \ref{fig:constant} shows the Y position of the robot in the scenario (b) with a fixed goal of -0.15 $\mathrm{m}$. 
As in simulation, the robot is able to track the reference value, due to the fact that the velocity takes into account the under-actuation of the trot gait.  
In this case, we decided to keep the reference generator always set to \textit{optimize} to demonstrate
that it is able to work properly also when the target has been reached, 
compensating drifts as in scenario (a).\\
In the last experiment, we show how we can use our reference generator
to react to external disturbances, see Fig. \ref{fig:screenshot}. 
As we have already mentioned in the Introduction, the task is not to reject the disturbance, but to cope with it and later recover from its effect.
An analysis of techniques to reject disturbances goes beyond the scope of this work. 
Figure \ref{fig:push} shows the Y position of the 
\gls{com} in a real hardware experiment in scenario (c) when the robot receives two manually applied pushes. The threshold on the error is set to 1
$\mathrm{cm}$ (dashed purple line).
During the push, the robot tries to resist to the disturbance and, thanks to the high-frequency 
re-planning of the \gls{nmpc}, it is able to keep the stability and avoid falling. 
Once the pushing force is removed, the reference generator drives the robot back towards the initial position.
As in the previous cases, the reference generator is set to \textit{optimize} when the robot is diverging from the goal.\\
The readers are encouraged to check the experiments 
corresponding to the mentioned results in the accompanying video.
\begin{figure}[h]
	\centering
	\includegraphics[scale = 0.25]{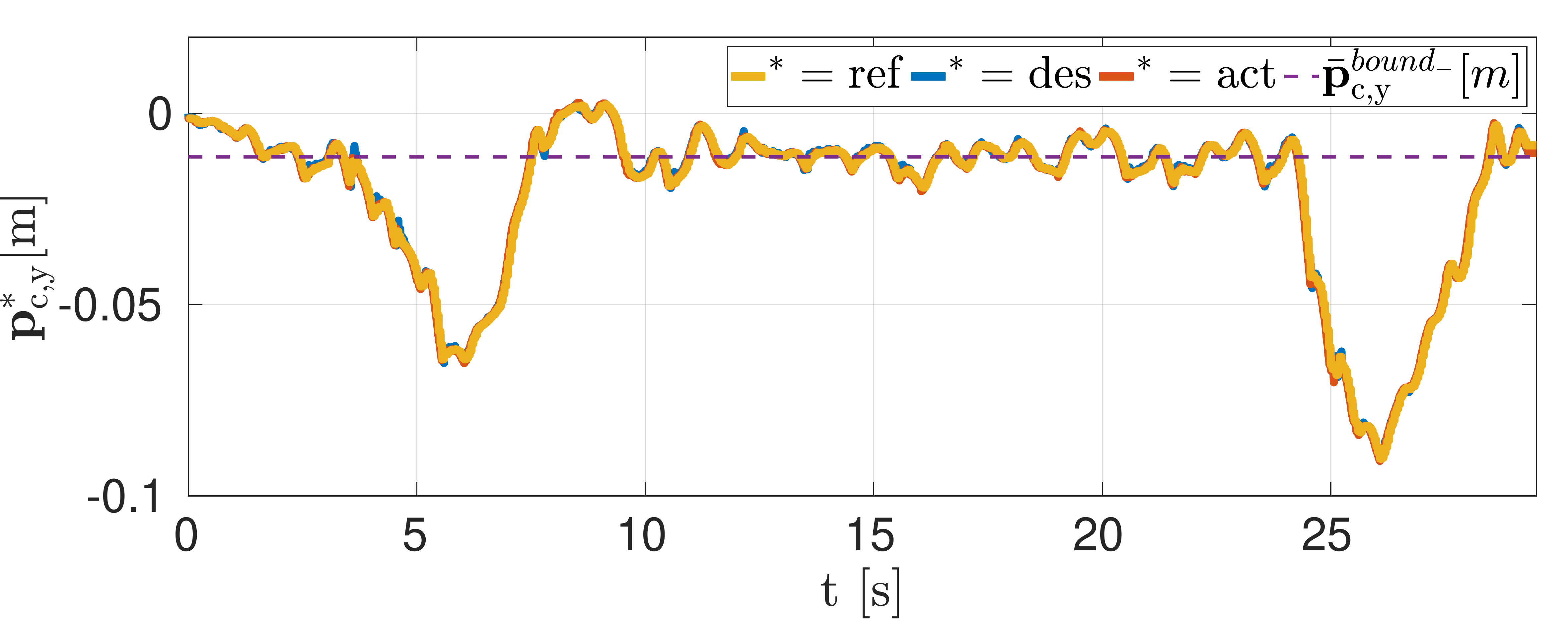} 
	\caption{Experiment, scenario (c): \gls{com} Y position of the robot. During the motion, the robot has been 
		pushed twice and it automatically comes back to the initial position when the push is removed.}.
	\label{fig:push}
\end{figure}
\section{Conclusions}
\label{sec:conclusion}
In this work, we presented a novel optimization-based reference generator for quadruped locomotion,
which deals with the under-actuation of statically unstable gaits and with external disturbances. 
It exploits the \gls{lip} model to compute feasible reference trajectories that allow the robot to accomplish a tracking task. 
A \gls{qp} mapping is used to determine the \gls{grfs} which correspond to the \gls{zmp} location computed by the \gls{lip} model. 
Velocities and \gls{grfs} are used as \textit{informative} tracking references by a 25 $\mathrm{Hz}$ lower-stage \gls{nmpc} planner introduced in \cite{openaccess}.
This results in the absence of conflicting tasks in the cost function, which simplifies the tuning of the cost weights of the \gls{nmpc}.
We validated our approach by performing simulations and experiments with the 22 $\mathrm{kg}$ quadruped robot Aliengo 
in three different scenarios: (a) straight motion, (b) fixed lateral goal and (c) recovery after a push.
For the last scenario, we demonstrated that the simple solution of adding a Cartesian 
\gls{pd} in parallel to the  \gls{nmpc} is not enough to return in a predefined time to the required position.
In addition, we presented and validated in simulations a formulation with slack variables that can guarantee 
to reach the goal in a specified time, without the need to further tune any parameters.\\
Future work involves extending the \textit{optimize} reference generator also to the 
heading (i.e., yaw) orientation, and to perform additional experiments with 
the real platform with the formulation that respects a specified response time.

\section*{Acknowledgements}
The publication was created with the co-financing of the European Union FSE-REACT-EU, PON Research and Innovation 2014-2020 DM1062 / 2021.\\
We thank prof. Mario Zanon and prof. Alberto Bemporad of IMT School for Advanced Studies Lucca for their precious advice and support.

\vspace{6pt} 
\reftitle{References}
\bibliography{root}

\begin{thebibliography}{999}

\bibitem[Rathod \em{et~al.}(2021)Rathod, Bratta, Focchi, Zanon, Villarreal,
  Semini, and Bemporad]{openaccess}
Rathod, N.; Bratta, A.; Focchi, M.; Zanon, M.; Villarreal, O.; Semini, C.;
  Bemporad, A.
\newblock Model Predictive Control With Environment Adaptation for Legged
  Locomotion.
\newblock {\em IEEE Access} {\bf 2021}, {\em 9},~145710--145727.
\newblock {\url{https://doi.org/10.1109/ACCESS.2021.3118957}}.

\bibitem[Raibert and Tello(1986)]{raibert1986legged}
Raibert, M.H.; Tello, E.R.
\newblock Legged Robots That Balance.
\newblock {\em IEEE Expert} {\bf 1986}, {\em 1}.
\newblock {\url{https://doi.org/10.1109/MEX.1986.4307016}}.

\bibitem[Winkler \em{et~al.}(2018)Winkler, Bellicoso, Hutter, and
  Buchli]{Winkler2018a}
Winkler, A.W.; Bellicoso, C.D.; Hutter, M.; Buchli, J.
\newblock {Gait and Trajectory Optimization for Legged Systems Through
  Phase-Based End-Effector Parameterization}.
\newblock {\em IEEE Robotics and Automation Letters (RA-L)} {\bf 2018}, {\em
  3},~1560--1567.
\newblock {\url{https://doi.org/10.1109/LRA.2018.2798285}}.

\bibitem[Bratta \em{et~al.}(2020)Bratta, Orsolino, Focchi, Barasuol, Muscolo,
  and Semini]{Bratta2020}
Bratta, A.; Orsolino, R.; Focchi, M.; Barasuol, V.; Muscolo, G.G.; Semini, C.
\newblock On the Hardware Feasibility of Nonlinear Trajectory Optimization for
  Legged Locomotion based on a Simplified Dynamics.
\newblock In Proceedings of the IEEE International Conference on Robotics and
  Automation (ICRA),  2020, pp. 1427--1423.
\newblock {\url{https://doi.org/10.1109/ICRA40945.2020.9196903}}.

\bibitem[Li and Wensing(2020)]{Li2020}
Li, H.; Wensing, P.M.
\newblock Hybrid Systems Differential Dynamic Programming for Whole-Body Motion
  Planning of Legged Robots.
\newblock {\em IEEE Robotics and Automation Letters (RA-L)} {\bf 2020}, {\em
  5}.
\newblock {\url{https://doi.org/10.1109/LRA.2020.3007475}}.

\bibitem[Semini \em{et~al.}(2011)Semini, Tsagarakis, Guglielmino, Focchi,
  Cannella, and Caldwell]{semini11hyqdesignjsce}
Semini, C.; Tsagarakis, N.G.; Guglielmino, E.; Focchi, M.; Cannella, F.;
  Caldwell, D.G.
\newblock Design of HyQ - a Hydraulically and Electrically Actuated Quadruped
  Robot.
\newblock {\em IMechE Part I: Journal of Systems and Control Engineering} {\bf
  2011}, {\em 225},~831--849.
\newblock {\url{https://doi.org/10.1177/0959651811402275}}.

\bibitem[Minniti \em{et~al.}(2022)Minniti, Grandia, Farshidian, and
  Hutter]{minniti22}
Minniti, M.V.; Grandia, R.; Farshidian, F.; Hutter, M.
\newblock Adaptive CLF-MPC With Application to Quadrupedal Robots.
\newblock {\em IEEE Robotics and Automation Letters (RA-L)} {\bf 2022}, {\em
  7},~565--572.
\newblock {\url{https://doi.org/10.1109/LRA.2021.3128697}}.

\bibitem[Hong \em{et~al.}(2020)Hong, Kim, and Park]{Hong}
Hong, S.; Kim, J.H.; Park, H.W.
\newblock Real-Time Constrained Nonlinear Model Predictive Control on SO(3) for
  Dynamic Legged Locomotion.
\newblock In Proceedings of the International Conference on Intelligent Robots
  and Systems (IROS),  2020, pp. 3982--3989.
\newblock {\url{https://doi.org/10.1109/IROS45743.2020.9341447}}.

\bibitem[Amatucci \em{et~al.}(2022)Amatucci, Kim, Hwangbo, and
  Park]{Amatucci2022}
Amatucci, L.; Kim, J.H.; Hwangbo, J.; Park, H.W.
\newblock Monte Carlo Tree Search Gait Planner for Non-Gaited Legged System
  Control.
\newblock In Proceedings of the IEEE International Conference on Robotics and
  Automation (ICRA),  2022, pp. 4701--4707.
\newblock {\url{https://doi.org/10.1109/ICRA46639.2022.9812421}}.

\bibitem[{Fahmi} \em{et~al.}(2019){Fahmi}, {Mastalli}, {Focchi}, and
  {Semini}]{Fahmi2019RAL}
{Fahmi}, S.; {Mastalli}, C.; {Focchi}, M.; {Semini}, C.
\newblock Passive Whole-Body Control for Quadruped Robots: Experimental
  Validation Over Challenging Terrain.
\newblock {\em {IEEE} Robotics and Automation Letters ({RA-L})} {\bf 2019},
  {\em 4},~2553--2560.
\newblock {\url{https://doi.org/10.1109/LRA.2019.2908502}}.

\bibitem[Meduri \em{et~al.}(2022)Meduri, Shah, Viereck, Khadiv, Havoutis, and
  Righetti]{Meduri2022}
Meduri, A.; Shah, P.; Viereck, J.; Khadiv, M.; Havoutis, I.; Righetti, L.
\newblock BiConMP: A Nonlinear Model Predictive Control Framework for Whole
  Body Motion Planning.
\newblock {\em IEEE Transaction on Robotics (T-RO)} {\bf 2022}.
\newblock accepted, {\url{https://doi.org/10.48550/ARXIV.2201.07601}}.

\bibitem[{Di Carlo} \em{et~al.}(2018){Di Carlo}, {Wensing}, {Katz}, {Bledt},
  and {Kim}]{DiCarlo2018}
{Di Carlo}, J.; {Wensing}, P.M.; {Katz}, B.; {Bledt}, G.; {Kim}, S.
\newblock Dynamic Locomotion in the MIT Cheetah 3 Through Convex
  Model-Predictive Control.
\newblock In Proceedings of the 2018 IEEE/RSJ International Conference on
  Intelligent Robots and Systems (IROS),  2018, pp. 1--9.
\newblock {\url{https://doi.org/10.1109/IROS.2018.8594448}}.

\bibitem[Grimminger \em{et~al.}(2020)Grimminger, Meduri, Khadiv, Viereck,
  Wuthrich, Naveau, Berenz, Heim, Widmaier, Flayols, Fiene, Badri-Sprowitz, and
  Righetti]{Grimminger2020}
Grimminger, F.; Meduri, A.; Khadiv, M.; Viereck, J.; Wuthrich, M.; Naveau, M.;
  Berenz, V.; Heim, S.; Widmaier, F.; Flayols, T.;  et~al.
\newblock An Open Torque-Controlled Modular Robot Architecture for Legged
  Locomotion Research.
\newblock {\em Robotics and Automation Letters (RA-L)} {\bf 2020}, {\em
  5},~3650--3657.
\newblock {\url{https://doi.org/10.1109/lra.2020.2976639}}.

\bibitem[Barasuol \em{et~al.}(2013)Barasuol, Buchli, Semini, Frigerio,
  De~Pieri, and Caldwell]{Barasuol2013}
Barasuol, V.; Buchli, J.; Semini, C.; Frigerio, M.; De~Pieri, E.R.; Caldwell,
  D.G.
\newblock A reactive controller framework for quadrupedal locomotion on
  challenging terrain.
\newblock In Proceedings of the IEEE International Conference on Robotics and
  Automation (ICRA),  2013, pp. 2554--2561.
\newblock {\url{https://doi.org/10.1109/ICRA.2013.6630926}}.

\bibitem[Cebe \em{et~al.}(2021)Cebe, Tiseo, Xin, Lin, Smith, and
  Mistry]{Cebe2020}
Cebe, O.; Tiseo, C.; Xin, G.; Lin, H.C.; Smith, J.; Mistry, M.N.
\newblock Online Dynamic Trajectory Optimization and Control for a Quadruped
  Robot.
\newblock {\em 2021 IEEE International Conference on Robotics and Automation
  (ICRA)} {\bf 2021}, pp. 12773--12779.
\newblock {\url{https://doi.org/10.1109/ICRA48506.2021.9561592}}.

\bibitem[Bouyarmane and Kheddar(2018)]{Bouyarmane18}
Bouyarmane, K.; Kheddar, A.
\newblock On Weight-Prioritized Multitask Control of Humanoid Robots.
\newblock {\em IEEE Transactions on Automatic Control} {\bf 2018}, {\em
  63},~1632--1647.
\newblock {\url{https://doi.org/10.1109/TAC.2017.2752085}}.

\bibitem[{Bledt} and {Kim}(2019)]{Bledt2019a}
{Bledt}, G.; {Kim}, S.
\newblock Implementing Regularized Predictive Control for Simultaneous
  Real-Time Footstep and Ground Reaction Force Optimization.
\newblock In Proceedings of the IEEE International Conference on Intelligent
  Robots and Systems (IROS),  2019, pp. 6316--6323.
\newblock {\url{https://doi.org/10.1109/IROS40897.2019.8968031}}.

\bibitem[Bjelonic \em{et~al.}(2022)Bjelonic, Grandia, Geilinger, Harley,
  Medeiros, Pajovic, Jelavic, Coros, and Hutter]{Bjelonic2022}
Bjelonic, M.; Grandia, R.; Geilinger, M.; Harley, O.; Medeiros, V.S.; Pajovic,
  V.; Jelavic, E.; Coros, S.; Hutter, M.
\newblock Offline motion libraries and online MPC for advanced mobility skills.
\newblock {\em The International Journal of Robotics Research} {\bf 2022}, {\em
  41},~903--924.
\newblock {\url{https://doi.org/https://doi.org/10.1177/02783649221102473}}.

\bibitem[Bemporad(1998)]{Bemporad98}
Bemporad, A.
\newblock Reference governor for constrained nonlinear systems.
\newblock {\em IEEE Transactions on Automatic Control} {\bf 1998}, {\em
  43},~415--419.
\newblock {\url{https://doi.org/10.1109/9.661611}}.

\bibitem[Kolmanovsky \em{et~al.}(2014)Kolmanovsky, Garone, and
  Di~Cairano]{Kolmanovsky14}
Kolmanovsky, I.; Garone, E.; Di~Cairano, S.
\newblock Reference and command governors: A tutorial on their theory and
  automotive applications.
\newblock In Proceedings of the 2014 American Control Conference,  2014, pp.
  226--241.
\newblock {\url{https://doi.org/10.1109/ACC.2014.6859176}}.

\bibitem[Garone \em{et~al.}(2017)Garone, Di~Cairano, and Kolmanovsky]{Garone17}
Garone, E.; Di~Cairano, S.; Kolmanovsky, I.
\newblock Reference and command governors for systems with constraints: A
  survey on theory and applications.
\newblock {\em Automatica} {\bf 2017}, {\em 75},~306--328.
\newblock
  {\url{https://doi.org/https://doi.org/10.1016/j.automatica.2016.08.013}}.

\bibitem[Kajita \em{et~al.}(2001)Kajita, Kanehiro, Kaneko, Yokoi, and
  Hirukawa]{Kajita01}
Kajita, S.; Kanehiro, F.; Kaneko, K.; Yokoi, K.; Hirukawa, H.
\newblock The 3D linear inverted pendulum mode: a simple modeling for a biped
  walking pattern generation.
\newblock In Proceedings of the International Conference on Intelligent Robots
  and Systems (IROS),  2001, pp. 239--246.
\newblock {\url{https://doi.org/10.1109/IROS.2001.973365}}.

\bibitem[Orin \em{et~al.}(2013)Orin, Goswami, and Lee]{Orin2013a}
Orin, D.E.; Goswami, A.; Lee, S.H.
\newblock {Centroidal dynamics of a humanoid robot}.
\newblock {\em Auton. Robots} {\bf 2013}, {\em 35}.
\newblock {\url{https://doi.org/10.1007/s10514-013-9341-4}}.

\bibitem[Mastalli \em{et~al.}(2022)Mastalli, Merkt, Xin, Shim, Mistry,
  Havoutis, and Vijayakumar]{Mastalli2022}
Mastalli, C.; Merkt, W.; Xin, G.; Shim, J.; Mistry, M.; Havoutis, I.;
  Vijayakumar, S.
\newblock Agile Maneuvers in Legged Robots: a Predictive Control Approach.
\newblock {\em arXiv} {\bf 2022}.

\bibitem[Focchi \em{et~al.}(2020)Focchi, Orsolino, Camurri, Barasuol, Mastalli,
  Caldwell, and Semini]{Focchi2020}
Focchi, M.; Orsolino, R.; Camurri, M.; Barasuol, V.; Mastalli, C.; Caldwell,
  D.G.; Semini, C.
\newblock Heuristic Planning for Rough Terrain Locomotion in Presence of
  External Disturbances and Variable Perception Quality.
\newblock {\em Springer Tracts in Advanced Robotics (STAR)} {\bf 2020}, pp.
  165--209.
\newblock {\url{https://doi.org/https://doi.org/10.1007/978-3-030-22327-4}}.

\bibitem[Nobili \em{et~al.}(2017)Nobili, Camurri, Barasuol, Focchi, Caldwell,
  Semini, and Fallon]{nobili_camurri2017rss}
Nobili, S.; Camurri, M.; Barasuol, V.; Focchi, M.; Caldwell, D.G.; Semini, C.;
  Fallon, M.
\newblock Heterogeneous Sensor Fusion for Accurate State Estimation of Dynamic
  Legged Robots.
\newblock In Proceedings of the Proceedings of Robotics: Science and Systems,
  2017.
\newblock {\url{https://doi.org/https://doi.org/10.15607/RSS.2017.XIII.007}}.

\bibitem[Durrant-Whyte and Bailey(2006)]{Durrant06}
Durrant-Whyte, H.; Bailey, T.
\newblock Simultaneous localization and mapping: part I.
\newblock {\em IEEE Robotics and Automation Magazine} {\bf 2006}, {\em
  13},~99--110.
\newblock {\url{https://doi.org/10.1109/MRA.2006.1638022}}.

\bibitem[Nowicki \em{et~al.}(2017)Nowicki, Belter, Kostusiak, Cížek, Faigl,
  and Skrzypczyński]{Belter16}
Nowicki, M.; Belter, D.; Kostusiak, A.; Cížek, P.; Faigl, J.; Skrzypczyński,
  P.
\newblock An experimental study on feature-based SLAM for multi-legged robots
  with RGB-D sensors.
\newblock {\em Industrial Robot} {\bf 2017}, {\em 44},~428--441.
\newblock {\url{https://doi.org/10.1108/IR-11-2016-0340}}.

\bibitem[Hult \em{et~al.}(2019)Hult, Zanon, Gros, and Falcone]{Hult19}
Hult, R.; Zanon, M.; Gros, S.; Falcone, P.
\newblock Optimal Coordination of Automated Vehicles at Intersections: Theory
  and Experiments.
\newblock {\em IEEE Transactions on Control Systems Technology} {\bf 2019},
  {\em 27},~2510--2525.
\newblock {\url{https://doi.org/10.1109/TCST.2018.2871397}}.

\bibitem[Chignoli and Wensing(2020)]{Chignoli20}
Chignoli, M.; Wensing, P.M.
\newblock Variational-Based Optimal Control of Underactuated Balancing for
  Dynamic Quadrupeds.
\newblock {\em IEEE Access} {\bf 2020}, {\em 8},~49785--49797.
\newblock {\url{https://doi.org/10.1109/ACCESS.2020.2980446}}.

\bibitem[Harada \em{et~al.}(2003)Harada, Kajita, Kaneko, and
  Hirukawa]{Harada2003}
Harada, K.; Kajita, S.; Kaneko, K.; Hirukawa, H.
\newblock ZMP analysis for arm/leg coordination.
\newblock In Proceedings of the International Conference on Intelligent Robots
  and Systems (IROS),  2003, pp. 75--81.
\newblock {\url{https://doi.org/10.1109/IROS.2003.1250608}}.

\bibitem[Vukobratovic and Borovac(2004)]{Vukobratovic04}
Vukobratovic, M.; Borovac, B.
\newblock Zero-Moment-Point - Thirty five years of its life.
\newblock {\em International Journal of Humanoid Robotics} {\bf 2004}, {\em
  01},~157--173.
\newblock {\url{https://doi.org/10.1142/S0219843604000083}}.

\bibitem[Bellicoso \em{et~al.}(2018)Bellicoso, Jenelten, Gehring, and
  Hutter]{Bellicoso18}
Bellicoso, C.D.; Jenelten, F.; Gehring, C.; Hutter, M.
\newblock Dynamic Locomotion Through Online Nonlinear Motion Optimization for
  Quadrupedal Robots.
\newblock {\em IEEE Robotics and Automation Letters (RA-L)} {\bf 2018}, {\em
  3},~2261--2268.
\newblock {\url{https://doi.org/10.1109/LRA.2018.2794620}}.

\bibitem[Frison and Diehl(2020)]{Frison2020}
Frison, G.; Diehl, M.
\newblock HPIPM: a high-performance quadratic programming framework for model
  predictive control.
\newblock {\em IFAC-PapersOnLine} {\bf 2020}, {\em 53},~6563--6569.
\newblock {\url{https://doi.org/10.1016/j.ifacol.2020.12.073}}.

\bibitem[Verschueren \em{et~al.}(2021)Verschueren, Frison, Kouzoupis, van
  Duijkeren, Zanelli, Novoselnik, Frey, Albin, Quirynen, and
  Diehl]{Verschueren2019}
Verschueren, R.; Frison, G.; Kouzoupis, D.; van Duijkeren, N.; Zanelli, A.;
  Novoselnik, B.; Frey, J.; Albin, T.; Quirynen, R.; Diehl, M.
\newblock {Acados - A modular open-source framework for fast embedded optimal
  control}.
\newblock {\em Math. Prog. Comp.} {\bf 2021}.
\newblock {\url{https://doi.org//10.1007/s12532-021-00208-8}}.

\bibitem[Guennebaud \em{et~al.}(2011)Guennebaud, Furfaro, and
  Gaspero]{Guennebaud2011}
Guennebaud, G.; Furfaro, A.; Gaspero, L.D.
\newblock eiquadprog.hh,  2011.

\end{thebibliography}
\end{document}